\documentclass[a4paper,11pt]{article}
\usepackage[utf8]{inputenc}
\usepackage{graphicx}
\usepackage{geometry}
\usepackage{array}
\usepackage{setspace}
\usepackage{url}
\usepackage{multirow}
\usepackage{footnote}
\usepackage{microtype}
\usepackage{graphicx}
\usepackage{subfigure}
\usepackage{amsmath}
\usepackage{graphics}
\usepackage{textcomp}

\usepackage{multicol}
\usepackage{color}
\usepackage{float}

\usepackage{booktabs}       
\usepackage{amsfonts}       
\usepackage{nicefrac}       
\usepackage{microtype}

\newcommand{\re}[1]{\textcolor{black}{#1}}

\begin{document}

\title{Machine Learning Applications on Neuroimaging for Diagnosis and Prognosis of Epilepsy: A Review}
\author{Jie Yuan$^1$, Xuming Ran$^1$, Keyin Liu$^1$, Chen Yao$^2$, \\
Yi Yao$^3$, Haiyan Wu$^4$, Quanying Liu$^{1\ast}$}
\date{}
\maketitle

\begin{centering}
$^1$ Shenzhen Key Laboratory of Smart Healthcare Engineering, Department of Biomedical Engineering, Southern University of Science and Technology, Shenzhen 518055, P.R. China
\\
$^2$ Shenzhen Second People’s Hospital, Shenzhen 518035, P.R. China
\\
$^3$ Shenzhen Children's Hospital, Shenzhen 518017, P.R. China
\\
$^4$ Centre for Cognitive and Brain Sciences and Department of Psychology, University of  Macau, Taipa, Macau
\\

\vspace{0.1cm}
$^\ast$ Corresponding author: \texttt{liuqy@sustech.edu.cn to Q.L.}
\\
\vspace{0.6cm}
\end{centering}

\setlength{\baselineskip}{20pt}

\begin{abstract}
Machine learning is playing an increasingly important role in medical image analysis, spawning new advances in the clinical application of neuroimaging. There have been some reviews on machine learning and epilepsy before, \re{and} they mainly focused on electrophysiological signals such as electroencephalography (EEG) and stereo electroencephalography (SEEG), while \re{neglecting} the potential of neuroimaging in epilepsy research. Neuroimaging has its important advantages in confirming the range of the epileptic region, \re{which is essential} in presurgical evaluation and assessment after surgery. However, \re{it is difficult for EEG} to locate the accurate epilepsy lesion region in the brain. In this review, we emphasize the interaction between neuroimaging and machine learning in the context of epilepsy diagnosis and prognosis. 
We start with an overview of \re{epilepsy and} typical neuroimaging modalities used in epilepsy clinics, MRI, DWI, fMRI, and PET.
Then, we \re{elaborate two} approaches in applying machine learning methods to neuroimaging data: i) the \re{conventional machine learning} approach combining \re{manual} feature engineering and classifiers, ii) the deep learning approach, such as the convolutional neural networks and autoencoders.
Subsequently, the application of machine learning on epilepsy neuroimaging, such as segmentation, localization, and lateralization tasks, as well as tasks directly related to diagnosis and prognosis are looked into in detail. Finally, we discuss the current achievements, challenges, and potential future directions in this field, hoping to pave the way for computer-aided diagnosis and prognosis of epilepsy.

\end{abstract}

\vspace{0.5cm}

\textbf{Highlights}
\begin{itemize}
\item{Machine learning plays an increasingly important role in epilepsy neuroimaging.}

\item{We introduce conventional machine learning methods and deep learning methods.}

\item{Segmentation, localization, lateralization, diagnosis and prognosis tasks are listed.}

\item{We discuss achievements and directions in machine learning for epilepsy neuroimaging.}
\end{itemize}

\section{Introduction}
Epilepsy is a neurological disease characterized by abnormal neurophysiological activity leading to epileptic seizures or abnormal behavior, accompanied by varying degrees of loss of sensation or consciousness. Unlike a single-source disease, epilepsy is usually associated with a set of chronic recurrent transient brain dysfunction syndromes~\cite{fisher2014ilae}. People with intractable epilepsy usually suffer from severe health problems and may lose the ability to take care of themselves. In 2017, The global lifetime epilepsy incidence rate was 7.60\textperthousand, which caused a huge global disease burden~\cite{fiest2017prevalence}. 
Therefore, the diagnosis and prognosis of epilepsy are important research topics. At present, advances in machine learning and neuroimaging have brought fresh air to this long-lasting research field.

The pathophysiological cause of epileptic seizures is the abnormal discharge of neurons, manifested as high-amplitude bursts on the Electroencephalogram (EEG). 
To clarify the concept, here we regard seizure as a transient brain dysfunction caused by excessive synchronous firing of neurons, and the epileptogenic foci are the sites of the epileptic attacks.
The detection and quantification of epileptogenic foci are essential for the diagnosis of epilepsy. Although about 70\% of patients with epilepsy can obtain effective seizure control through anti-epileptic drugs~\cite{eadie2012shortcomings}, the remaining 30\% of patients have failed seizure control, leading to drug-resistant epilepsy or intractable epilepsy.  Intractable epilepsy has a high mortality rate and a poor prognosis, requiring surgical treatment~\cite{fisher2014ilae}.
Surgical treatment of epilepsy can be divided into two categories, palliative surgery, and radical surgery, according to whether the surgery is targeted the epileptic foci~\cite{schramm2008surgery}.
\textit{Palliative surgery} (\textit{i.e.} corpus callosotomy or neuromodulation) aims at the seizure-related neural circuits, rather than directly at the epileptogenic foci~\cite{spencer1989corpus}, while \textit{radical surgery} (\textit{i.e.} radiofrequency thermocoagulation, resection, and dissection) directly deals with epileptogenic foci~\cite{clusmann2002prognostic}. All operations require precisely locating the epileptogenic foci, and neuroimaging modalities are the doctor's main diagnostic tool. Neuroimaging has its important advantages in confirming the range and size of epileptogenic foci compared to the EEG, which means a lot in presurgical evaluation and assessment after surgery. Therefore, there is a great need for automated analysis of neuroimaging to help clinicians.

\begin{figure}[t]
\centering
\includegraphics[width=0.8\textwidth]{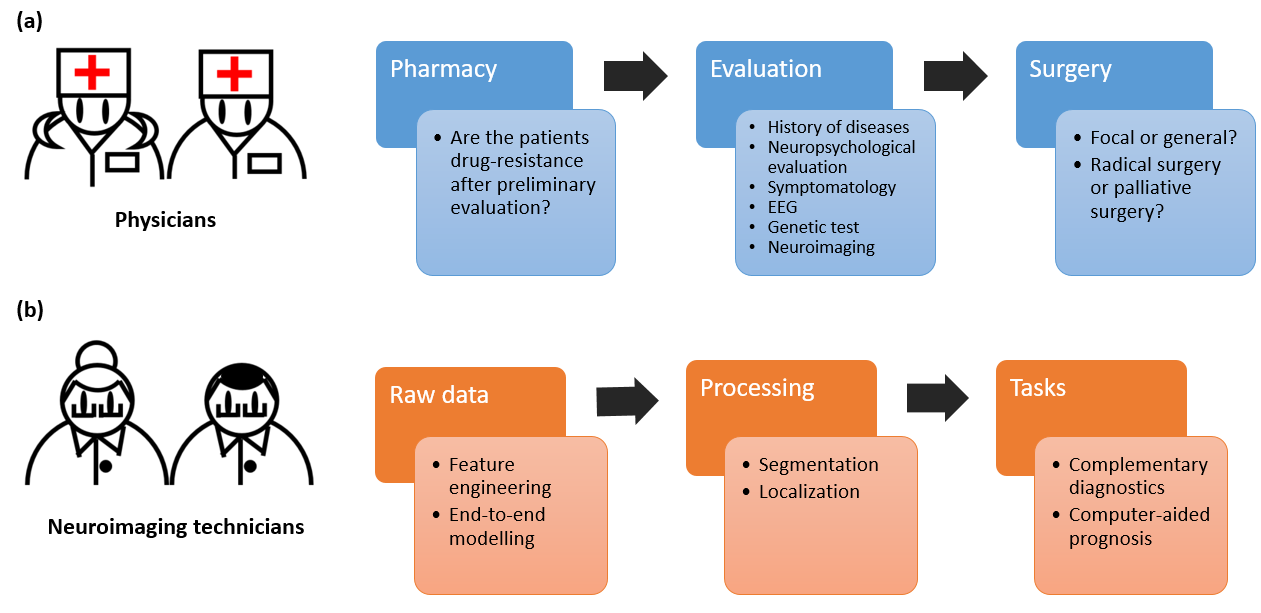}
\caption{Epilepsy diagnosis workflow for clinicians and neuroimaging technicians. (a) Clinicians initially treat the semeiology-diagnosed epilepsy patients with medication. If the patient is drug-resistance, the clinician will make a presurgical evaluation and multimodal images will be collected. Then, the clinician decides the type of surgery, either radical surgery or palliative surgery, based on whether the epilepsy is focal or general. (b) Neuroimaging technicians analyze medical data to support clinicians. They extract features from raw data via feature engineering or directly train end-to-end models for segmentation or positioning. Their ultimate tasks are computer-aided diagnosis and prognosis.}
\label{fig:workflow}
\end{figure}

The clinical workflow of epilepsy diagnosis and presurgical evaluation is illustrated in \textbf{Figure~\ref{fig:workflow}(a)}. Patients with suspected epilepsy are first screened by non-invasive techniques for diagnosis, and then those who are diagnosed with epilepsy are usually recommended to take anti-epileptic drugs. For patients with intractable epilepsy, clinicians have to conduct the further comprehensive evaluation, including locating the epileptogenic foci and judging whether they contain the \textit{eloquent cortex}, which -- if removed -- will result in loss of sensory processing or linguistic ability, or paralysis. This routine evaluation procedure before surgical treatment is called \textit{presurgical evaluation}~\cite{juhasz2020utility, zijlmans2019changing}.
Palliative surgery would be suggested for intractable epilepsy patients whose epileptogenic foci involve the eloquent cortex, while radical surgery is considered for those whose epileptogenic foci are not in the eloquent cortex. \re{In the latter case, clinicians need to locate the epileptic source accurately in order to preserve the largest functional eloquent cortex}~\cite{nunes2012diagnosis}. Therefore, more advanced and even invasive screening techniques might be involved.  
\re{On the one hand, epileptic seizures can be directly located by electrophysiological techniques, such as non-invasive electroencephalograph (EEG) and invasive electrocorticography (ECoG)~\cite{west2019surgery}. On the other hand, brain lesions, presumably indirectly or directly leading to clinical seizures, can be detected by neuroimaging techniques, such as magnetic resonance image (MRI)~\cite{zhao2017role} and positron emission tomography (PET)~\cite{la2009pet}.}
In general, when the epileptogenic foci are identifiable in neuroimaging, the chance of no seizures after radical surgery increases by about 2-3 times~\cite{tellez2010surgical}.

However, the clinical workflow is laborious. The outcome it finally achieved with all those medical data is sometimes full of uncertainty because of the eye fatigue of clinicians~\cite{shoeibi2021applications}. Due to the importance and difficulty of presurgical evaluation, technicians are often expected to help clinicians analyze various neuroimaging data in the epilepsy department. The workflow for neuroimaging technicians is shown in \textbf{Figure~\ref{fig:workflow}(b)}. Their ultimate tasks are complementary diagnosis and computer-aided prognosis. 
Voxel-based morphometry (VBM) methods are the traditional computer-aided ways \re{for technicians to analyze neuroimaging}~\cite{keller2008voxel}. Voxel-wise statistical comparisons between \re{normalized normal and abnormal} brain images are always needed. However, VBM methods only capitalize on superficial information of brain images \re{such as volume and thickness} and sometimes \re{cause misclassification between images} due to improper registration. Advances in machine learning methods have made it possible to analyze images in depth.

Previous reviews mainly focused on the machine learning applied to electrophysiological data (such as EEG) for epilepsy ~\cite{acharya2019characterization, boonyakitanont2020review, shoeibi2021epileptic}, very few reviews have paid attention to epilepsy neuroimaging so far~\cite{abbasi2019machine, kini2016computational}. In this review, other than electrophysiology, we emphasize the role of neuroimaging in epilepsy and the application of machine learning to epilepsy neuroimaging. We will introduce the neuroimaging techniques and machine learning models used in epilepsy study, to highlight the potentials of the machine learning applications in computer-aided diagnosis and prognosis.

We searched Pubmed, Scopus, and Google scholar for papers with the keywords `machine learning', `deep learning', `epilepsy', `CT', `MRI', `SPECT' and `PET' in the title or abstract. We surveyed more than 120 papers and checked the citations of them. We then excluded articles that used epilepsy datasets for super-resolution or other non-clinical related tasks, or \re{those} that were ancient. The latest update to the included papers is on February 11, 2021.

The rest of this review is structured as follows. We first present the neuroimaging modalities commonly used in epilepsy (\textbf{Section 2}). Then we briefly introduce the \re{methodological background}, including the \re{conventional machine learning  approaches and deep learning} approaches (\textbf{Section 3}). We then present \re{various} machine learning tasks for epilepsy and review previous studies on each task (\textbf{Section 4}). Finally, we discuss the achievements and challenges in the field to call for more attention on the machine learning on neuroimaging for epilepsy (\textbf{Section 5}).

\begin{figure}[H]
\centering
\includegraphics[width=0.8\textwidth]{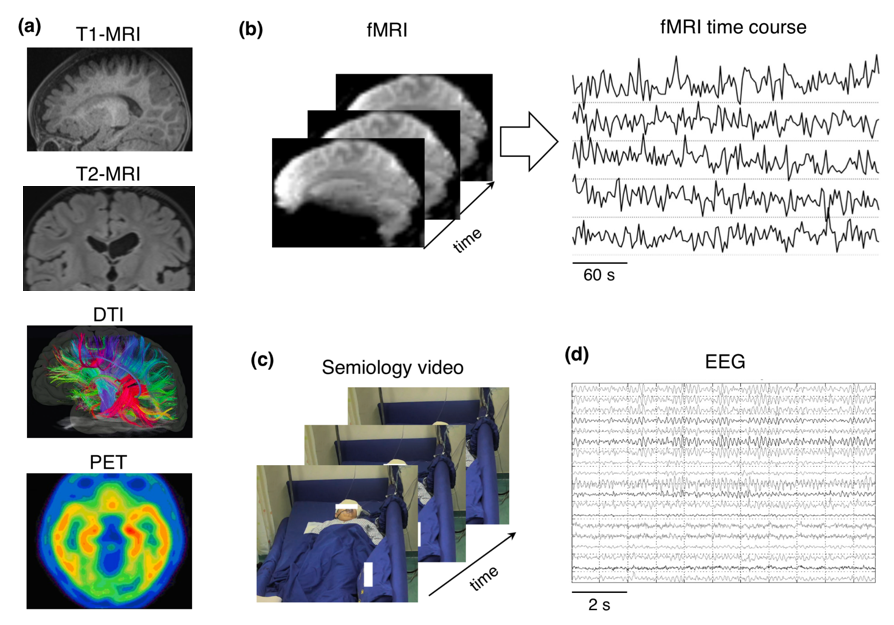}
\caption{The non-invasive multi-modal images and electrophysiology for diagnosis and prognosis of epilepsy, including (a) T1, T2, DTI, PET images, (b) fMRI, (c) semiology video, (d) EEG.} 
\label{fig:modality}
\end{figure}

\section{Neuroimaging tools for epilepsy}

The development of medical imaging technology, especially neuroimaging, opens a window for studying the brain through imaging the structure and examining the functional dynamics. The widely-used image modalities are listed in \textbf{Figure~\ref{fig:modality}}. Various non-invasive neuroimaging techniques can be used to monitor brain structure and function, including T1-weighted MRI (T1w MRI), T2-weighted MRI (T2w MRI), diffusion tensor imaging (DTI), functional MRI (fMRI), and PET.

A typical MRI system consists of the following \re{hardware and software} components: a magnet, a gradient coil, \re{a radiofrequency} transmitting coil, a radiofrequency receiving coil, and signal processing and image reconstruction \re{algorithms}. The nuclear spin of a hydrogen atom in human body can be equivalent to a small magnetic dipole. In a strong magnetic field, the radiofrequency transmitting coil flips the hydrogen nuclei from the direction of the main field to the transverse plane, and their differences form a net magnetization vector. The hydrogen nuclei precess around the main magnetic field. The gradient coil generates magnetic fields whose strength varies with spatial position, and these magnetic fields are used for spatial coding of the signals. \re{The induced current signal is received by the radiofrequency receiving coil and recorded, and eventually the image can be reconstructed by signal processing and image reconstruction algorithms.} Fine-tuning the parameters (\re{\textit{e.g.} the} flip angle or \re{the} pulse interval) lead to a variety of MRI sequences, such as T1w, T2w, and T2w-FLAIR MRI, and each sequence highlights different tissues of the brain\re{, such as gray matter or white matter}~\cite{giedd2004structural}.
Specifically, T1w MRI maps the anatomical structure of the brain; T2w MRI captures the aberrant zone in the white matter; T2w-FLAIR MRI provides high contrasts between the gray matter and cerebrospinal fluid (CSF). \re{These neuroimages can aid} the detection of patients with focal cortical dysplasia (FCD)~\cite{adler2017novel}. For example, it has been reported that the combination of conventional visual examination and morphometric MRI analysis has significantly high diagnostic sensitivity in both subgroups of FCD (94\% for FCD \uppercase\expandafter{\romannumeral2}a; 99\% for FCD \uppercase\expandafter{\romannumeral2}b)~\cite{wagner2011morphometric}.

Unlike the traditional structural MRI, diffusion imaging leverages the extent, directionality, and organization of the motion of free water to provide image contrast~\cite{colombo2009imaging}. Diffusion-weighted imaging (DWI) detects water \re{molecules} diffusion through the transverse magnetization direction, resulting in a phase shift caused by signal attenuation. As an improvement of DWI, DTI reflects the direction of white matter fiber bundles. Diffusion kurtosis imaging (DKI) depicts the molecular weight of water that diffuses out of the Gaussian distribution in the tissue. 
The Kurtosis information reflects the non-Gaussian characteristics caused by the complex structure of multiple microcellular compartments~\cite{steven2014diffusion}.
Fractional anisotropy (FA) and mean diffusion (MD) maps are commonly used parameter maps derived from DTI and DKI, while the mean kurtosis (MK) map is merely obtained from DKI. In particular, FA measures the degree of directionality, while MD measures the average diffusion along all diffusion directions. MK describes a more complex spatial distribution, that is, the average of the diffusion kurtosis along all diffusion directions. They are important imaging biomarkers for the detection of heterogeneous samples~\cite{arab2018principles}.

The brain imaging technique has been extended from structure to function, and some imaging systems based on brain function have been designed, including functional MRI (fMRI)~\cite{kwong1992dynamic} and PET~\cite{sweet1955localization}. Magnetic resonance imaging is used to detect the changes in cerebral blood flow and metabolism caused by the neural activity because of the paramagnetism of blood. Thus fMRI can reflect the activities of the brain tissues. PET utilizes radiotracers to observe the local uptake-related variation, which could reflect the abnormal metabolism in brain. For instance, as a complement, PET can \re{locate} the regional interictal hypometabolism, guide neuroradiologists to look for lesions~\cite{velez2012neuroimaging}.


Although EEG directly reflects the abnormal neural activities in epilepsy and thus more appealing for epilepsy detection, \re{the potential of} neuroimaging is largely underestimated in epilepsy applications.  In \re{this review}, we will mainly focus on neuroimaging in epilepsy-related machine learning tasks.

\section{Machine learning methods \re{applied in epilepsy}}

Machine learning, as a data-geared method, builds a myriad of mathematical models that can learn from the structured training data to make predictions or decisions in novel contexts and the newly-presented data, without being \re{explicitly programmed} to perform that task~\cite{bishop2006pattern}.
A wide range of machine learning models have been applied to neuroimaging to aid the diagnosis and prognosis of epilepsy~\cite{wang2018multimodal, cantor2015detection, alaverdyan2020regularized, farazi2017lateralization, memarian2015multimodal, gleichgerrcht2018deep}. 
\re{As the use of machine learning in epilepsy grows}, it is necessary to review and summarize methods, tasks, scientific findings, and their interpretations. Here, we categorize the machine learning methods into \re{two} classes \re{in the context of neuroimaging applications}: \re{i) the conventional machine learning approach and ii) the deep learning approach}. \re{The conventional machine learning approach} mostly performs hand-craft feature engineering, following a classification or a regression task. The \re{deep learning} approach applies deep neural networks \re{to a specific task or to extract features automatically}. 

There are various open-source tools for machine learning and especially for processing neuroimaging of epilepsy. Firstly, Matlab, Scikit-learn, Keras, TensorFlow, PyTorch, Caffe, and Theano are some well-known toolkits to implement machine learning models. Secondly, the codes for  neuroimaging processing related to epilepsy are available on the Github. \footnote{Epileptic lesion detection: \url{https://github.com/MELDProject/MELDProject.github.io} }\footnote{Focal cortical dysplasia detection: \url{https://github.com/kwagstyl/FCDdetection/}}\footnote{Rs-fMRI alignment to cortical stimulation: \url{https://github.com/jarodroland/Peds_rfMRI_vs_ECS}}\footnote{Hippocampus segmentation on epilepsy: \url{https://github.com/MICLab-Unicamp/e2dhipseg}}

\begin{figure}[H]
\centering
\includegraphics[width=0.75\textwidth]{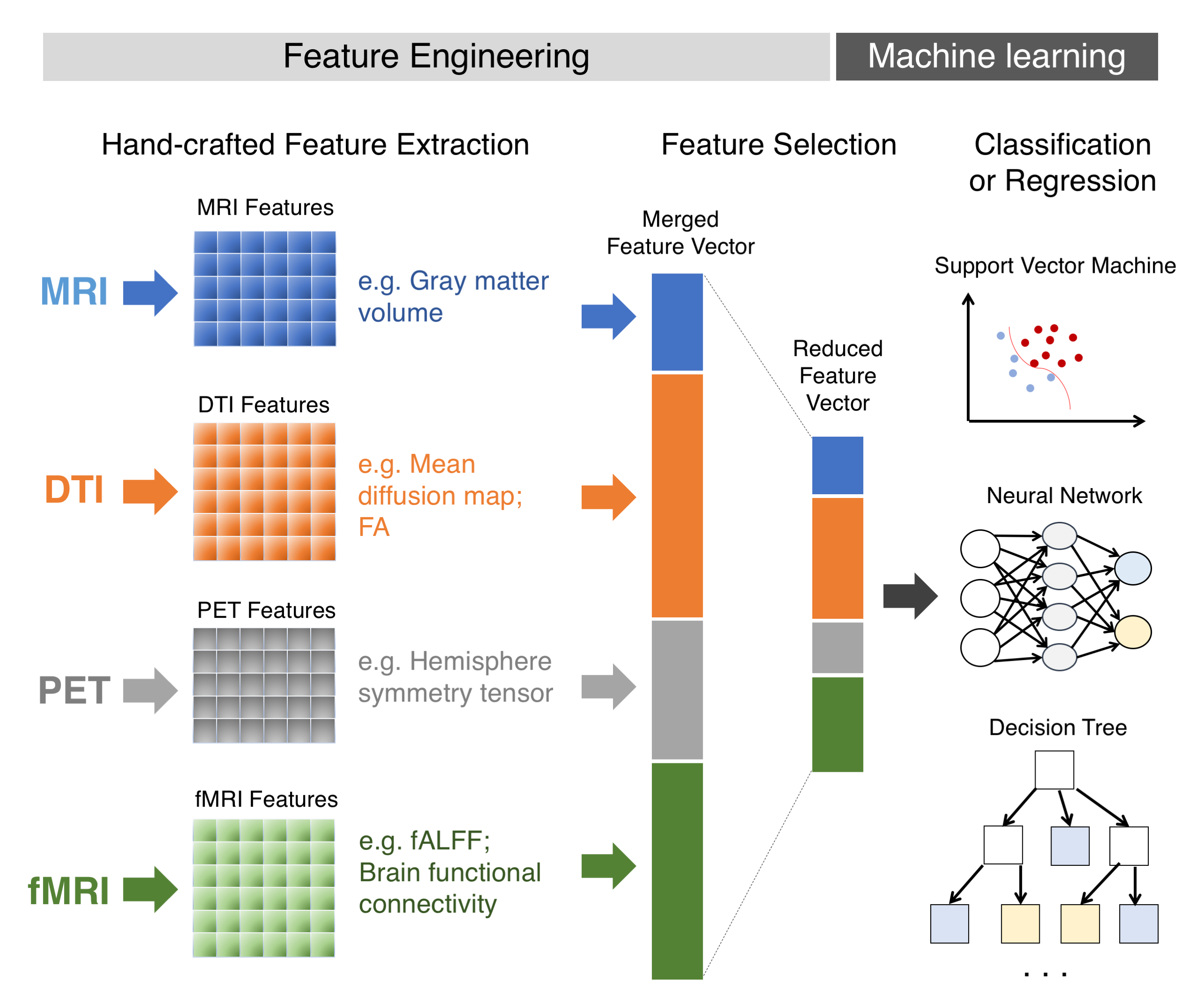}
\caption{The \re{conventional machine learning} approach. It consists of a feature engineering step to extract and select features from single/multiple neuroimaging modalities, and a machine learning step to perform a classification or regression task.} 
\label{fig:composition}
\end{figure}

\subsection{Conventional machine learning approach}

The \re{conventional machine learning} approach consists of two steps, a manually feature engineering step and a machine learning step (\textbf{Figure \ref{fig:composition}}). The feature engineering step extracts the hand-crafted features from brain images. The machine learning step then inputs those features to a machine learning model for a certain task, such as classification (to detect the impaired or normal brain)~\cite{alaverdyan2020regularized} or regression (to predict the severity of epilepsy)~\cite{munsell2019relationship}. The machine learning model applied in \re{this} approach is usually a simple classifier, rather than the deep neural network.

\subsubsection{Feature engineering}

Feature engineering is a necessary step in the conventional machine learning models. Extracting the distinctive features from the medical images \re{can effectively reduce the data dimension} and \re{prevent a model from overfitting}.
In \textbf{Table~\ref{tab:features}}, we \re{list} some representative features extracted from each neuroimaging modality relevant to epilepsy. 

\begin{table}[]
\caption{Representative hand-crafted features used in epilepsy}
\vspace{-5mm}
\begin{center}
\small
\begin{tabular}{m{5em}m{37em}m{2em}}
\\[-1.8ex]\hline 
\hline \\[-1.8ex] 
\textbf{Modality} & \textbf{Features} & \textbf{Ref} \\
\hline
\multirow{3}{*}{T1w MRI} & Mean, standard deviation, variance, energy, and entropy of segmented  hippocampus &  ~\cite{sayed2021characterization}\\
\cline{2-3}
& Cortical thickness, intensity at the grey-white matter contrast, curvature, sulcal depth, intrinsic curvature & ~\cite{wagstyl2020planning} \\
\hline
\multirow{2}{*}{T2w-MRI} & Volume and intensity sampled on the medial sheet of hippocampus & ~\cite{kim2014multivariate}\\
\cline{2-3}
& First-order statistical and volumetric gray-level co-occurrence matrix texture features & ~\cite{sahebzamani2019machine}\\
\hline
\multirow{2}{*}{FLAIR-MRI} & Intensity sampled at 25\%, 50\% and 75\% of the cortical thickness and at the grey-white matter boundary & ~\cite{wagstyl2020planning}\\
\cline{2-3}
& Image intensity features and wavelet-based texture features & ~\cite{jafari2010flair}\\
\hline
DTI & Mean diffusion and Fractional Anisotropy & ~\cite{del2017using}\\
\hline
\multirow{2}{*}{DKI} & Mean Kurtosis & ~\cite{del2017using}\\
\cline{2-3}
& MD, FA, MK and the fusion of FA and MK & ~\cite{huang2020identifying}\\
\hline
\multirow{2}{*}{fMRI} & Fractional amplitude of low-frequency fluctuation (fALFF) & ~\cite{wang2018multimodal}\\
\cline{2-3}
& BOLD signal values in different regions, hemispheres and tasks & ~\cite{torlay2017machine}\\
\hline
PET & Hemisphere symmetry tensor & ~\cite{jiang2019transfer}\\
\hline
\end{tabular}

\centering
\label{tab:features}
\end{center}

\footnotesize{\tiny \textit{Abbrev}: MD, Mean diffusion; FA, Fractional Anisotropy; MK, Mean Kurtosis; BOLD, blood oxygenation level-dependent.}\\
\end{table}

The extracted features are usually with large redundancy as we would like to maintain as much information as possible in the original medical images. Thus, feature selection for removing the invalid features and distilling the relevant features is our next step. The dimension reduction techniques, such as Principal components analysis (PCA), have been widely used to transfer data from the original high-dimensional space into a low-dimensional space with minimal information loss by selecting the most distinguishable features or create new features~\cite{motoda2002feature}.  
PCA aims at mapping the high-dimensional features to a low-dimensional space through linear projections via maximizing the variance of 
the matrix data on the projected dimension. \re{It works best when the variance distribution in different dimensions of the data set is uneven. The principal components of the feature matrix are obtained by sorting the eigenvectors of its covariance matrix according to the value of the corresponding eigenvalues.} In this way, a lower-dimensional representation (\textit{i.e.} the principal components) of features contains the intrinsic characteristics of the original dataset.

After feature engineering, the selected features will be input into the machine learning models for real-world tasks. To be noted, the performance of conventional machine learning models, such as linear discrimination analysis and support vector machine, relies heavily on the extracted features. 

\subsubsection{Linear discrimination analysis}
Linear discrimination analysis (LDA, also called Fisher linear discrimination analysis) is a classic supervised method in machine learning problems such as data dimensionality reduction, feature extraction, and pattern recognition. LDA was first described in a two-class problem by Ronald A. Fisher~\cite{https://doi.org/10.1111/j.1469-1809.1936.tb02137.x}, and later generalized as multi-class linear discriminant analysis. Different from the variance maximization theory of PCA, the main idea of LDA is to project the data from a high-dimensional space onto a lower-dimensional space so that the same classes are clustered together while the different classes are far apart. Firstly, the mean of class $i$ is given by:
\begin{equation}
u_{i}=\frac{1}{n_{i}} \sum_{x \in \text { class }i} x
\label{equ:meanofclass}
\end{equation}
The overall sample mean is given by:
\begin{equation}
u=\frac{1}{m} \sum_{i=1}^{m} x_{i}
\label{equ:overallmean}
\end{equation}

According to the definition of the between-class scatter matrix $S_b$ and the within-class scatter matrix $S_w$, the following formula can be obtained:
\begin{equation}
\mathbf{S}_{b}=\sum_{i=1}^{c} \boldsymbol{n_i}\left(\boldsymbol{u}_{i}-\boldsymbol{u}\right)\left(\boldsymbol{u}_{i}-\boldsymbol{u}\right)^{T}
\label{equ:betweenclassscatter}
\end{equation}
\begin{equation}
\mathbf{S}_{w}=\sum_{i=1}^{c} \sum_{x_{k} \in \text { class } i}\left(u_{i}-\boldsymbol{x}_{k}\right)\left(\boldsymbol{u}_{i}-\boldsymbol{x}_{k}\right)^{T}
\label{equ:withinclassscatter}
\end{equation}

\re{By maximizing the following objective functions, the within-class distance is minimized and the between-class distance is maximized: }
\begin{equation}
J_{fisher}(w)=\frac{w^{T} S_{b} w}{w^{T} S_{w} w}= \frac{\sum\limits_{i=1}^{c} n_{i} w^{T}\left(u_{i}-u\right)\left(u_{i}-u\right)^{T} w}{\sum\limits_{i=1}^{c}  \sum\limits_{x_{k} \in \text { class }i} w^{T}\left(u_{i}-x_{k}\right)\left(u_{i}-x_{k}\right)^{T} w}
\label{equ:fisher}
\end{equation}
where the optimal transformation $w$ would maximize the objective function.

LDA is a pre-processing step in Machine Learning and pattern classification \re{applications}. 
The variants of LDA include quadratic discriminant analysis (QDA)~\cite{McLachlan1992DiscriminantAA}, flexible discriminant analysis (FDA)~\cite{10.2307/2290989}, and regularized discriminant analysis (RDA)~\cite{friedman1989regularized}.

\subsubsection{Random forest}
Random Forest (RF) is a combination of decision trees. Each tree depends on the value of an independently sampled random vector and has the same distribution for all trees in the forest~\cite{10.1023/A:1010933404324}. RF creates Bagging ensemble~\cite{breiman1996bagging} based on decision tree learner, and further introduces random \re{features} selection in the training process of decision tree. Specifically, the traditional decision tree selects an optimal \re{one} from the \re{feature} set of the current nodes when selecting \re{features}. However, for RF, a subset containing $k$ \re{features} is randomly selected from the \re{feature} set of the nodes in the decision tree, and then an optimal \re{feature} is selected from this subset for division. 

\subsubsection{Support vector machines}
\label{svm}

\re{Initial} SVM aims to find a hyperplane in $N$-dimensional space to separate the samples from two classes, where $N$ is the number of features~\cite{cortes1995support}.
\re{It was originally} designed for two-class classification but \re{was later} extended to multi-class classification. \re{Besides,} support vector regression (SVR)~\cite{Smola2004ATO} is known for solving regression problems. 

The SVM problem for the training data set of $m$ points $
\left\{x_{i}, y_{i}\right\}_{i=1}^{m}$
 is given by:
\begin{equation}
\begin{aligned}
  \min \frac{1}{2}\|w\|^{2}+C \sum_{i=1}^{m}\left(\xi_{i}+\xi_{i}^{*}\right) \\
  s.t. \:  \left\{\begin{array}{l}y_{i}-\left(w^{T} x_{i}+b\right) \leq \varepsilon+\xi_{i} \\ \left(w^{T} x_{i}+b\right)-y_{i} \leq \varepsilon+\xi_{i}^{*} \\ \xi_{i}, \xi_{i}^{*} \geq 0\end{array}\right.
\end{aligned}
\label{equ:svmobjectivefunc}
\end{equation}
\re{where $w$ and $b$ are the learning parameters corresponding to the weight vector and the bias.} They determine a hyperplane, 
$f\left(x_{i}\right)=w^{T} x_{i}+b$, \re{which can separate two classes}. $C$ as the regularization parameter and $\epsilon$ is a margin of tolerance. \re{These hyperparameters have to be} pre-defined in order to solving the optimization problem. 
$\xi_{i}$ and $\xi_{i}^{*}$ are the slack variables~\cite{Smola2004ATO}.

The kernel function of SVM is to take data as input and transform it into the required form and define it as follows:
\begin{equation}
K(\bar{\mathbf{x}})=\left\{ \begin{array}{l}1, \ \text { if }\|\bar{\mathbf{x}}\| \leq 1 \\ 0, \ \text { otherwise }\end{array} \right.
\label{equ:kerneldefine}
\end{equation}
where the value of this function is 1 inside the closed ball of radius 1 centered at the origin, and 0 otherwise.

\re{A variety of} kernel functions \re{have been proposed for SVMs}. 
Gaussian kernel is a general-purpose kernel. It is usually used when no prior knowledge about the data is known. \re{The} equation of Gaussian kernel is:
\begin{equation}
k(\mathbf{x}, \mathbf{y})=\exp \left(-\frac{\|\mathbf{x}-\mathbf{y}\|^{2}}{2 \sigma^{2}}\right)
\label{equ:gauskernel}
\end{equation}
The Gaussian radial basis function (RBF) is defined as:
\begin{equation}
k\left(\mathbf{x}_{\mathbf{i}}, \mathbf{x}_{\mathbf{j}}\right)=\exp \left(-\gamma\left\|\mathbf{x}_{\mathbf{i}}-\mathbf{x}_{\mathbf{j}}\right\|^{2}\right)
\label{equ:rbfkernel}
\end{equation}
where $\gamma>0$.
The polynomial kernel \re{is defined} as follow:
\begin{equation}
k\left(\mathbf{x}_{\mathbf{i}}, \mathbf{x}_{\mathbf{j}}\right)=\left(\mathbf{x}_{\mathbf{i}} \cdot \mathbf{x}_{\mathbf{j}}+1\right)^{d}
\label{equ:polykernel}
\end{equation}
where $d$ is the degree of the polynomial kernel. \re{The polynomial kernel is popular in image processing.}

\re{Many} extensions and variants of SVMs have been proposed \re{for the versatility of SVMs}, such as, one-class SVM (OC-SVM)~\cite{10.5555/3009657.3009740}, least square SVM (LS-SVM)~\cite{Suykens2004LeastSS,Suykens2002WeightedLS,Liu2016RampLL}, fuzzy SVM~\cite{Lin2002FuzzySV,Lin2005FuzzySV,Lin2004TrainingAF}, weighted SVM~\cite{Yang2005WeightedSV}, transductive SVM~\cite{Cevikalp2017LargescaleRT}, and twin SVM~\cite{Jayadeva2007TwinSV}. \re{They significantly facilitate the classification of neuroimaging.}

\subsubsection{Shallow neural networks}
\label{snn}
Neural network refers to the machine learning model \re{with} a layered network architecture \re{which is composed of many artificial neurons in each layer.}  The \re{artificial} neurons are connected between layers, and the connection strength is the learning parameter \re{(i.e. weight)}. A large number of studies have reported that neural networks can achieve superior performance in complex medical tasks~\cite{2020An}. 
The shallow neural network has formed the groundwork for the deep neural network (DNN)~\cite{hring2020expressivity}, which can be trained rapidly and flexibly with error backpropagation on large data sets~\cite{Rumelhart1986LearningRB}. 

Shallow neural networks usually use a fully-connected architecture, \re{which consists} of neurons, weights, and bias. Neurons have three types: input unit, hidden unit, and output unit.
When the input vector is fed into the network as an input unit, it will propagate through the network to the output unit, \re{and it represents the probability of a particular category.} Additionally, the expression of each neuron in hidden units is as follows:

\begin{equation}
    y\;=\;\sigma\left(\sum_{i=1}^nw_ix_i\;-\;b\right).
\label{equ:neuron}
\end{equation}
\re{where} $w_1,...,w_n$ are the weights and $b$ is the bias. Also, \re{$\sigma(\cdot)$} is the activation function that provides a non-linear element, enabling the shallow neural network suffices to approximate any well-behaved functions like bounded continuous functions~\cite{billings1992properties}.

The \re{choice of} loss function depends on the specific task, for example, the cross-entropy loss for classification tasks, and the root mean squared error (RMSE) loss for regression tasks. \re{Sometimes} regularization terms are \re{added into the loss function to represent our prior knowledge. For instance, the L1 norm ensures sparsity and L2 norm for smoothness}. The design of loss function is still an ongoing research topic in machine learning field. 

After the network structure and the loss function are designed, parameters of neural networks, such as weights and bias, can be learned by error backpropagation~\cite{Rumelhart1986LearningRB}.


\subsection{Deep learning approach}
The \re{deep learning approach} trains a learning system with multiple functional layers~\cite{Glasmachers2017LimitsOE} like the deep neural network with complex nonlinear mapping. Comparing to the \re{conventional machine learning} approach, the merit of the \re{deep learning model} is that the raw data can be directly input into the model without the need for complicated preprocessing operations on the neuroimaging. Notwithstanding the convenience of the end-to-end models, they have some limitations such as the demand for giant data size and the weak interpretability~\cite{guo2016deep}. 

\begin{figure}[H]
\centering
\includegraphics[width=0.9\textwidth]{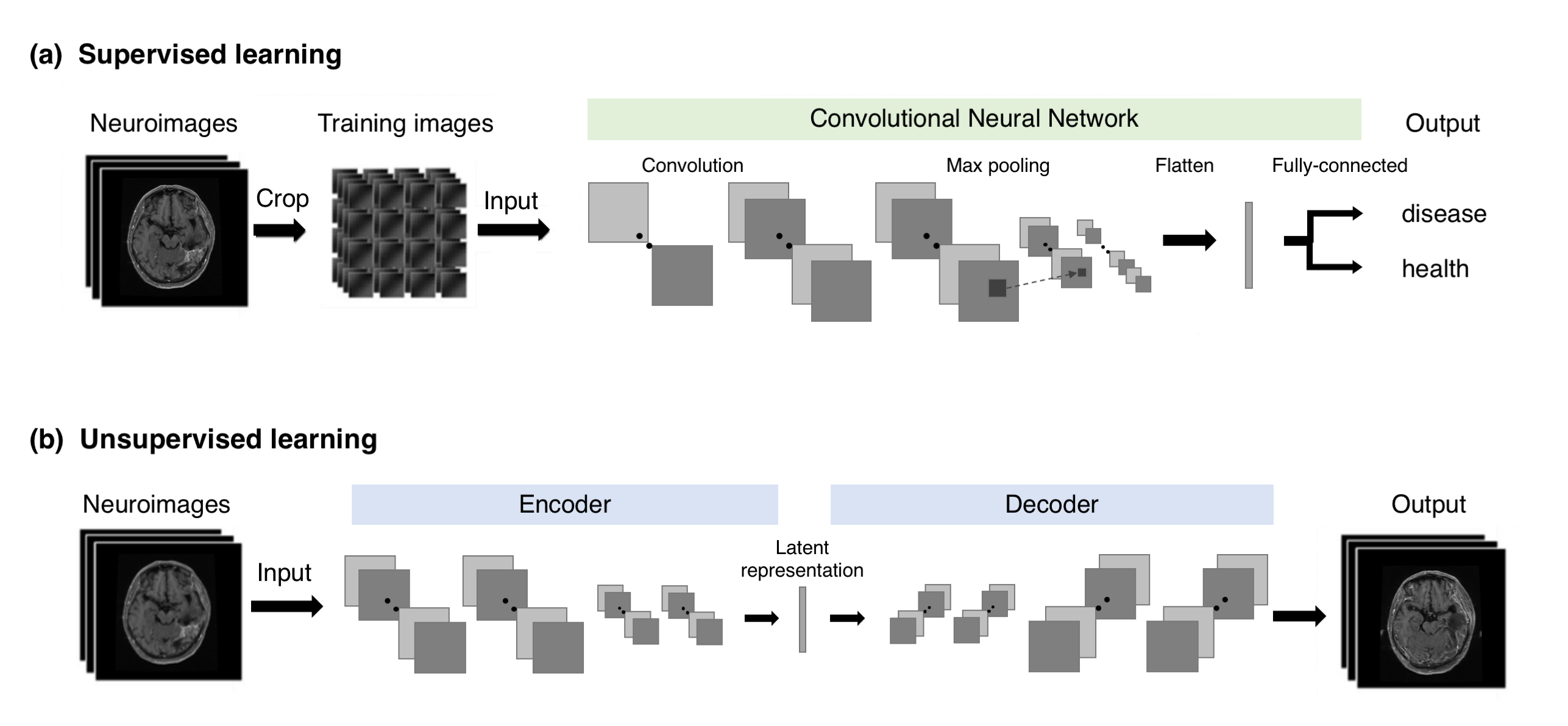}
\caption{The \re{deep learning approach}. Examples of a supervised learning model (a) and an unsupervised learning model (b) used for neuroimaging analysis.} 
\label{fig:end2end}
\end{figure}

\re{Deep learning models} for diagnosis and prognosis of epilepsy include the supervised learning models and the unsupervised learning models (\textbf{Figure~\ref{fig:end2end}}). Supervised learning models require labels (such as disease or health, the degree of severity, the types of epilepsy). \re{They can be trained to reveal a relationship between the features of data and the label, hoping to be generalized to the newly generated data without its label}~\cite{mohri2018foundations}. Unsupervised learning models do not require labels and can capture the patterns of probability densities or neuronal predilections by the intrinsic characteristics of features~\cite{hinton1999unsupervised}. 

\subsubsection{Convolutional neural networks}
\label{cnn}
Convolutional neural network (CNN)~\cite{LeCun1998GradientbasedLA,Krizhevsky2017ImageNetCW} is the most popular machine learning method nowadays \re{and} has been widely used in image processing. \re{The powerful ability of CNN to extract complex hidden features from high-dimensional data with deep convolutional structure enables it to be used as a feature extractor} in medical image classification~\cite{pominova2018voxelwise} and segmentation~\cite{kumar2018u,carmo2021hippocampus}.
CNN usually consists of a series of layers with each layer following a differentiable activation function. A variety of activation functions have been proposed, such as sigmoid, tanh, softmax, Rectified Linear Unit (ReLU)~\cite{Hahnloser2000DigitalSA,Nair2010RectifiedLU}, leaky ReLU~\cite{Krizhevsky2017ImageNetCW}, etc. \textbf{Figure \ref{fig:end2end}(a)} illustrates a typical CNN architecture with three types of neural layers: convolutional layer, pooling layer, and fully-connected layer. The convolutional layers are interspersed with pooling layers \re{and ended with} the fully connected layers. The convolutional layer takes a small patch of the input images (\textit{i.e.} 3 $\times$ 3 $\times$ 3 \re{for three-dimension image}), called the local receptive field, and then utilizes various learnable kernels to convolve the receptive field to generate multiple feature maps. A pooling layer performs the non-linear downsampling to reduce the input volume's spatial dimensions for the next convolutional layer \re{and to adapt the larger size features}. The fully-connected layer pools the 2D feature maps into a 1D feature vector. The local response normalization is a non-trainable layer and performs “lateral inhibition” by normalizing over the local input regions.

In practice, \re{one of the main problems} in training deep models is the over-fitting, \re{which is caused by} the gap between a limited number of training samples and a large number of learnable parameters. Overfitting during training would reduce the performance on the test dataset. Therefore, many approaches focused on avoiding overfitting such as dropout and batch normalization. Dropout is a typical deep learning technique, referring to randomly dropping a fraction of the units or connections during each training iteration. It has been \re{proven} that dropout can considerably avoid over-fitting~\cite{Srivastava2014DropoutAS}. \re{Batch normalization is performed as an additional regularization through the running average of the mean variance statistics in each mini-batch.} Batch normalization can drastically \re{accelerate the convergence speed of} training and improve the generalization performance~\cite{Ioffe2015BatchNA}

\re{Epilepsy diagnosis and prognosis} based on the CNN framework is currently an emerging and active research field~\cite{PICCIALLI2021111}. More details will be presented in the Section \ref{sec:applications}.

\subsubsection{Autoencoders}
\label{ae}

Auto-encoder (AE) is a typical unsupervised learning method, consisting of an encoder and a decoder (see \textbf{Figure \ref{fig:end2end}(b)}). \re{AEs learn the latent representation of input data and then reconstruct the input data into output.} AE involves multiple hidden layers which are stacked to form a deep network, which called deep auto-encoder (DAE). Compared with the shallow networks, DAEs are capable of discovering more complex patterns inherent in the input data due to the deep layers. So far, a number of AE variations have been proposed, such as denoising auto-encoder (denoising AE)~\cite{Vincent2008ExtractingAC}, sparse auto-encoder (sparse AE)~\cite{Ranzato2006EfficientLO}, variational auto-encoder (VAE)~\cite{Kingma2014AutoEncodingVB,Rezende2014StochasticBA}, adversarial auto-encoder (AAE)~\cite{Makhzani2015AdversarialA}, and stacked sparse auto-encoder (SSAE)~\cite{Shin2013StackedAF}. These extensions of AE have the potential \re{to learn} useful latent representations from neuroimaging data and \re{improve} the robustness of medical image reconstruction, which could serve better applications for epilepsy diagnosis. For example, stacked convolutional autoencoder could be used to learn \re{the representation of brain images}~\cite{alaverdyan2020regularized}.

\subsubsection{Hybrid models}

More recently, the hybrid approach combining \re{conventional machine learning and deep learning} has been brought into the epilepsy study~\cite{alaverdyan2020regularized,jiang2019transfer}. This \re{architecture} has high flexibility to improve the performance. 

The hybrid approach \re{starts with} extracting the hidden features as latent representations \re{through unsupervised model instead of} the hand-crafted features, and then \re{employs} machine learning algorithms for classification. For instance, convolutional autoencoders were stacked to form the siamese network \re{to learn} the representations of each patch at the same space in healthy brain MRIs. In the testing stage, the representations can further distinguish \re{the normal patches and} the abnormal patches with FCD lesions~\cite{alaverdyan2020regularized}. 
Besides the unsupervised end-to-end model, supervised deep neural networks are also capable of hidden feature extraction. As an example, Jiang et al. integrated four sets of features that were extracted from four classic deep neural networks (ResNet - 50, VGGNet - 16, Inception - V3, SVGG - C3D) and classified \re{the epileptic and the normal through these fused features} by a fully-connected  layer~\cite{jiang2019transfer}.

\section{Machine learning tasks}
\label{sec:applications}
Machine learning model learns the internal representation of multiple dimensions of numerical features \re{through} classification or regression. Then the new-come instance will be identified by its representation. Researchers have exploited multiple machine learning applications based on different objectives. \re{Specifically, some focused on segmentation, localization, and lateralization which belong to the neuroimaging processing tasks (\textbf{See section4.1}).
Some focused on the computer-aided diagnosis tasks (\textbf{See section4.2}). 
Finally, others emphasized the computer-aided prognosis tasks (\textbf{See section4.3}).}

\re{Different tasks have different evaluation criteria for the quality of the model used. Generally, the most basic evaluation criterion in machine learning tasks is accuracy (ACC), that is, the proportion of correctly classified samples to the total number of samples. In medical application, sensitivity and specificity are other common criterion.
Sensitivity refers to the proportion of patients whose lesions are accurately detected in all patients. Specificity is the proportion of control instances that are not classified into the patients among the controls. High sensitivity is equivalent to low missed diagnosis rate while specificity is equivalent to low misdiagnosis rate. Ideally, we want both sensitivity and specificity to be high, but in fact we find a balance between sensitivity and specificity. This process can be represented by the receiver operating characteristic (ROC) curve. The abscissa value of the curve is 1-specificity, and the ordinate value is sensitivity. After the curve is drawn, the area under ROC curve (AUC) can be calculated, which refers to the probability that the model ranks a random positive 
instance more highly than a random negative instance. The larger the value of AUC, the better the performance of the model. Besides, \textit{Dice coefficient} (\textit{i.e.} Dice similarity coefficient), as a measure of the similarity between two sets of data, is the most broadly used \re{assessment} to evaluate the performance of segmentation algorithms. The higher the dice coefficient is, the better the performance is.}

\begin{figure}[H]
\centering
\includegraphics[width=0.8\textwidth]{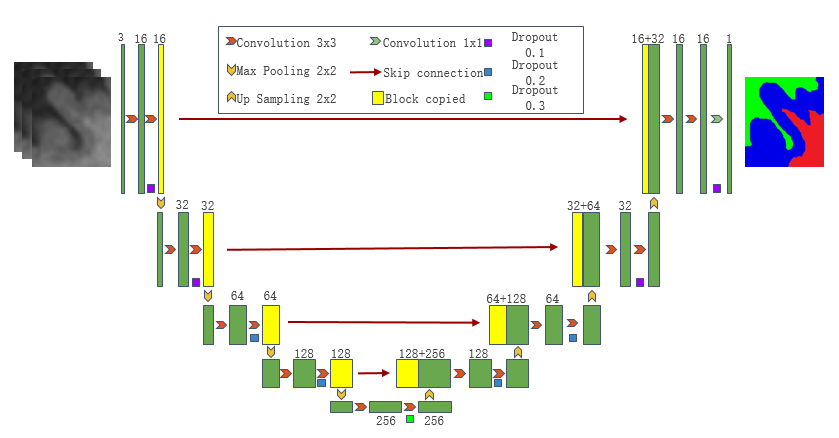}
\caption{An example of U-net for brain tissues segmentation. This framework is adapted from an open-source project~\cite{MLVisuals}.}
\label{fig:Unet}
\end{figure}

\subsection{Brain image processing}
Neuroimaging processing tasks are \re{for certain brain} areas including tissues that are in the fixed areas of all brains and abnormalities that are different from the ones in normal brains. They sometimes output the masks.
Among the methods, U-net~\cite{ronneberger2015u}, named for its network shape, is a commonly used tool for neuroimaging processing. It \re{specifically generates} masks in segmentation and localization tasks for epilepsy applications~\cite{carmo2021hippocampus,Dev2019AutomaticDA}. Additionally, the structure of U-net is shown in (\textbf{Figure~\ref{fig:Unet}})~\cite{MLVisuals}. U-net consists of an encoder to learn the representation of the image and a decoder to reconstruct the mask corresponding to original image. The encoder and decoder can be any deep neural network backbones, which 
enables U-net to adjust the structure according to the application.

\subsubsection{Segmentation}
The segmentation task generates a mask of the target region (\textit{i.e.} the region of interests) on a given brain image. This mask is usually output by a supervised deep learning model. So the model requires a prior label mask. In the training stage, the annotation of the mask requires cumbersome label work by clinical experts.  

The deep learning approach has been applied to \re{epilepsy-specific} regions.
Onofrey et al. applied a registration method \re{based on dictionary learning} to segment the post-surgical brain surface from the fusion of CT and MRI data. Specifically, \re{preoperative} MRI was used to guide the surgical resection of epileptic tissue, \re{and SEEG records use postoperative CT images as the standard for epileptic areas}. The dice coefficient between the segmented result and the standard was 94\%~\cite{onofrey2018segmenting}.

Also, segmentation of certain functional regions, especially the hippocampus, is of special interests in epilepsy.
The abnormalities in volume and shape of hippocampus have been reported as biomarkers for diagnosis of epilepsy~\cite{thom2014hippocampal}. 
\re{There was an end-to-end model trained on the epilepsy dataset to segment the hippocampus.}
Carmo et al. designed a CNN model based on the U-net architecture to obtain the segmentation mask of hippocampus, with the dice coefficient being 77\%. A native database \textit{HCUnicamp} was used in their study, which was collected by personnel from the Brazilian Institute of Neuroscience and Neurotechnology (BRAINN), containing 132 patients with epilepsy~\cite{carmo2021hippocampus}.



\subsubsection{Lesion Localization}
Applying to the localization tasks, machine learning models generate a mask to indicate the aberrant region. Some of the representative work are listed in \textbf{Table \ref{tab:localization}}.
Among various localization tasks, the localization of \re{cortical lesions} is one of the most important tasks. \re{Precise} localization of the pathological lesion will greatly inform \re{neurosurgeons}; in particular, \re{intractable epilepsy is usually associated with brain lesion.} Benefiting from advances in the field of computer vision, \re{various} deep learning methods have been proposed and effectively applied to the task of localizing epileptogenic lesions.

The field of FCD localization has aroused special interest of researchers due to its high incidence and misdiagnosis rate. 
The subtle lesions could be manifested on MRI as \re{thickened cortex, blurred gray matter-white matter interface, etc}~\cite{taylor1971focal}. \re{However, some FCD lesions may be seen as normal by clinicians on structural MRI when the histology is confirmed to be positive.} Studies \re{on this task usually use sensitivity and specificity as evaluation indicators and they
would define} the overlap standard between the localization outcome and ground truth.

\re{Conventional machine learning} methods have been widely used in this field. The earliest compositional method selected six feature maps to locate the clusters of candidate lesion voxels based on the one-class SVM via ranking T1w MRI voxels by size and suspicion degree~\cite{el2013computer}. This method \re{unbiasedly} detects \re{FCD lesions in two patients}~\cite{el2013computer}. A \re{follow-up} study by the same team captured more distinguishing features and reached general results to locate the FCD lesions. In three MRI-positive cases, the detection rate was 100\%, while in 10 MRI-negative cases, the detection rate was 70\%~\cite{el2016detection}. Further, the tree-structured hierarchical conditional random field was utilized to identify the \re{abnormally} segmented cortical surface regions patch-wise by thresholding the posterior probabilities, and it located the FCD region of MRI-positive \re{instances} with 90\% accuracy and MRI-negative \re{instances} with 80\% accuracy~\cite{ahmed2014hierarchical}. The accuracy of MRI-negative cases was the highest in the FCD localization field~\cite{ahmed2014hierarchical}. 

Three studies chose the shallow neural network as the classifier in \re{conventional machine learning method}~\cite{adler2017novel,jin2018automated,wagstyl2020planning}. With regard to \re{feature extraction}, FreeSurfer (a brain image processing software) is a commonly adopted feature extracted tool \footnote{\url{http://surfer.nmr.mgh.harvard.edu/}} for the \re{conventional} models.
Adler et al. added a novel feature (local cortical deformation) to the FreeSurfer's features, \re{which was designed for} the developing cortex of children. Sensitivity improved from 53\% to 73\% \re{with the novel features}~\cite{adler2017novel}. Jin et al. combined the surface-based morphometry feature sets extracted by FreeSurfer from T1w, T2w, and FLAIR MRI, effectively localizing the FCD with 73.7\% sensitivity and 90\% specificity~\cite{jin2018automated}.
Additionally, the morphological and intensity features could be extracted to locate the FCD regions with up to 73.5\% sensitivity and 100\% specificity by wagsty et al.~\cite{wagstyl2020planning}.

More recently, \re{deep learning} models were performed in the \re{field of FCD localization}.
Gill et al. leveraged MRI vertex and voxel respectively to \re{locate} the FCD region by decision tree (DT) and CNN~\cite{gill2017automated, gill2018deep}, among which CNN outperformed compositional method with 91\% sensitivity and 95\% specificity.
Furthermore, the customized U-Net was used to \re{locate} the FCD lesions using only FLAIR MRI in 2019 with 82.5\% accuracy~\cite{Dev2019AutomaticDA}.

Besides the \re{structural MRI}, diffusion images can identify the white matter tracks involved in epileptic abnormality. Xu et al. designed a CNN with a combined focal and central loss and a soft attention scheme, and applied it to \re{DWI streamline} data, achieving a clinically acceptable accuracy of 73\%-100\% to detect the white matter pathway connected to eloquent areas~\cite{xu2019objective}. A following-up study from the same group applied the CNN method to the DWI to \re{locate} the pediatric eloquent cortex ~\cite{lee2020novel}. In addition, MD map and the fusion of FA and MK maps extracted from DKI were segmented into hippocampus masks to perform abnormal identification of certain brain regions. This work can indirectly locate anomalies in the hippocampus. They classified the patients with hippocampus epilepsy and the healthy subjects through a deep CNN to \re{locate} the abnormal hippocampus with the accuracy of 90.8\%~\cite{huang2020identifying}.

FMRI during a task or at resting state can pinpoint the abnormal regions involved in a functional network, such as the language network. Torlay applied the Extreme Gradient Boosting algorithm (XGBoost) on fMRI data to identify the atypical patterns of language networks on a phonological and semantic language task in patients with focal epilepsy~\cite{torlay2017machine}. \re{Besides, using MLP to assign the resting state network to every voxel on the resting-state fMRI (rs-fMRI)} has been reported to \re{locate} the eloquent function like sensorimotor cortex~\cite{roland2019comparison}, which is essential in the presurgical evaluation.

\begin{table}[]
\caption{Localization of abnormal brain regions or pathways}
\vspace{-5mm}
\begin{center}
\small

\begin{tabular}{m{9em}m{9em}m{6em}m{9em}m{2em}m{2em}}
\\[-1.8ex]\hline 
\hline \\[-1.8ex] 
\re{\textbf{Task}} & \textbf{Inputs} & \textbf{Model} & \textbf{Performance} & \textbf{Data Size}$^*$ & \textbf{Ref} \\
\hline

\re{Locate} FCD region & MRI vertex & Intensity \&  RUSBoosted/ AdaBoosted DTs & Sensitivity: 83.0\%\newline{}Specificity: 92.0\% & 41\textbackslash{}38 & ~\cite{gill2017automated}\\
\hline
\re{Locate} FCD region & MRI voxel & CNN   & Sensitivity: 91.0\%\newline{}Specificity: 95.0\% & 107\textbackslash{}38 & ~\cite{gill2018deep}\\
\hline
\re{Locate} FCD region & Two texture parameters on T1-weighted brain MRI &  oc-svm & MRI-positive ACC: 100.0\%\newline{}MRI-negative ACC: 70.0\% & 11\textbackslash{}77 & ~\cite{el2016detection}\\
\hline
\re{Locate} FCD region & Cortical thickness, Gray/white-matter contrast, Sulcal depth, Curvature and Jacobian distortion & Restricted boltzmann machines + a bayesian non-parametric mixture model & MRI-positive ACC: 99.0\%\newline{} MRI-negative ACC: 58.0\% & 24\textbackslash{}50 & ~\cite{zhao2016non}\\
\hline
\re{Locate} FCD region & Morphological and intensity features & ANN   & Sensitivity: 73.7\%\newline{}Specificity: 90.0\% & 15\textbackslash{}35 & ~\cite{jin2018automated}\\
\hline
\re{Locate} FCD region & Morphological and intensity features & ANN   & Sensitivity: 73.5\%\newline{}Specificity: 100.0\% & 34\textbackslash{}38 & ~\cite{wagstyl2020planning}\\
\hline
\re{Locate} the atypical patterns of language networks & Atypical language patterns from fMRI & XGBoost & AUC: 91.5\% & 16\textbackslash{}39 & ~\cite{torlay2017machine}\\
\hline
\re{Locate} the sensorimotor network & rs-fMRI voxel & \re{Multilayer} perceptron   & Mean dice coefficient: nearly 50.0\% & 16 & ~\cite{roland2019comparison}\\
\hline
\re{Locate} the functionally white matter pathway including eloquent areas & DWI streamline & CNN   & ACC: 73.0\%-100.0\% & 70\textbackslash{}70 & ~\cite{xu2019objective}\\
\hline
\re{Locate} the functionally white matter pathway including eloquent areas & DWI tract segmentation & CNN   & ACC: 98.0\% & 89 & ~\cite{lee2020novel}\\
\hline
\re{Locate} hippocampus lesions & DKI hippocampus segmentation & CNN   & ACC: 90.8\% & 59\textbackslash{}70 & ~\cite{huang2020identifying}\\

\hline

\end{tabular}%
\centering

\label{tab:localization}
\end{center}

\footnotesize{\tiny$^*$ In $N_e$\textbackslash{}$N_c$, $N_e$ is the number of the patients while $N_c$ is the number of the controls.}\\
\end{table}

\subsubsection{Lateralization}
Lateralization is an important task that involves two different purposes: \re{1)} to identify the hemisphere where the epileptogenicity is \re{located}, \re{2)} is to identify the \re{dominant} hemisphere of the eloquent cortex, such as language regions. The former is helpful for lesion localization and epilepsy diagnosis. For example, the hippocampal atrophy usually lateralizes toward the focus in drug-resistant temporal lobe epilepsy (TLE). The latter is essential for the surgical \re{evaluation} because the eloquent cortex must be maintained during the treatment. Some results are listed in \textbf{Table \ref{tab:tasks-lateralization}}. 

The structural images have been widely used in lateralization tasks. For instance, Kim et al. manipulated LDA to classify the left- or right-sided seizure focus with manually extracted features after the segmentation of hippocampus in MRI~\cite{kim2014multivariate}. Similarly, the LDA classifier can be replaced by SVM or DNN classifier for higher accuracy~\cite{hosseini2014support, farazi2017lateralization}.
SVM \re{based on the graph theory network characteristics of the} DTI structural connectomes has been used to \re{classify the right TLE, the left TLE, and the control group}~\cite{kamiya2015machine}. Another study showed that even in visual MRI‐negative TLE, \re{it is possible to predict the laterality of epileptic seizures by training a RF classifier, which contains an ensemble of 5000 decision trees, using a set of 117 features from their structural brain images}~\cite{bennett2019learning}.

Besides the structural images, the functional images can also be utilized to lateralize the dominant hemisphere. Yang et al. extracted features from rs-fMRI within local brain regions, between brain regions, and across the whole network. Then the RF was used to reduce the dimension of data and SVM was employed to classify the laterality of TLE patients with 83\% accuracy~\cite{yang2015lateralization}. Gazit et al. suggested using the selected features in the verb-generation fMRI task, and then choosing the probabilistic logistic regression method \re{to help} lateralize the language regions of patients with epilepsy~\cite{gazit2016probabilistic}.

\begin{table}[]
\caption{\re{Tasks} of the lateralization}
\vspace{-5mm}
\begin{center}
\small
\begin{tabular}{m{8em}m{11em}m{6em}m{7em}m{3em}m{2em}}
\\[-1.8ex]\hline 
\hline \\[-1.8ex] 
\textbf{Tasks} & \textbf{Inputs} & \textbf{Model} & \textbf{Performance} & \textbf{Data Size} & \textbf{Ref} \\
\hline
\re{Lateralize epileptic foci} & Intensity-based vector MR features of hippocampal subfields  & LDA   & ACC: 93.0\% & 15 & \cite{kim2014multivariate}\\
\hline
\re{Lateralize epileptic foci} &
Shape features from the hippocampi of TLE & Ensemble of SVM, KNN, DT & ACC : 94.0\% & 40\textbackslash{}15 & \cite{farazi2017lateralization} \\
\hline
\re{Lateralize epileptic foci} & Volume and intensity-based manual features  & RF & AUC of MRI-positive: 98.1\%\newline{}AUC of MRI-negative: 84.2\% & 104 & \cite{bennett2019learning}\\
\hline
\re{Lateralize language regions} & \re{language lateralization index} & \re{logistic regression} & \re{ACC : 89\%} & \re{76} & \re{\cite{gazit2016probabilistic}} \\
\hline

\end{tabular}%
\centering
\label{tab:tasks-lateralization}
\end{center}
\end{table}

\subsection{Epilepsy Diagnosis}
Researchers characterize the \re{potential} epileptic subjects and reveal the essentiality of the features \re{by machine learning models}, which can guide the clinicians and improve the diagnostic evaluation workflow. On the one hand, these tasks were \re{designed} to save unnecessary time for well-trained doctors. On the other hand, algorithms with high sensitivity increase the chance of detecting potential 'invisible' regions that may be ignored by human experts. We listed some work relevant to epilepsy diagnosis in \textbf{Table~\ref{tab:tasks-classification}}.

The good old-fashioned \re{machine learning} methods had been widely applied to classify the patients and the healthy controls~\cite{cantor2015detection,wang2018multimodal,Jiang2017Hemisphere,del2017using,liedlgruber2019can,holler2020prediction}. It may also identify the most \re{contributory} features corresponding to a specific disease in the process.
\re{Some studies relied on several manual features through their prior knowledge.} For example, SVM was utilized to distinguish patients with tonic-clonic seizures from the normal cohort by using two hand-crafted MRI features (Gray matter volume from T1 and fALFF from fMRI). The fALFF feature got better performance with an accuracy of 83.72\%~\cite{wang2018multimodal}. Moreover, taking the DTI \re{features} (\textit{i.e.}, MD and FA) and the DKI feature (\textit{i.e.}, MK) into consideration, MK achieved the best accuracy (82\%) among three features based on SVM classifier~\cite{del2017using}.
Similar \re{conventional} methods have also been applied on PET images. \re{Features from the hemisphere symmetry tensor were extracted by multi-linear PCA}, and then SVM were applied to classify the abnormal images and the normal ones~\cite{Jiang2017Hemisphere}.\re{Other studies used a large number of features of neuroimaging data that can be easily extracted by existing toolbox (FreeSurfer) in feature engineering.} Usually, PCA was used to reduce the dimension of features, and then SVM can gain more discernible results with 88.90\% accuracy~\cite{cantor2015detection}.


\re{Deep learning} models have been brought into epilepsy diagnosis since 2018~\cite{pominova2018voxelwise}.
The first application of CNN for the recognition of epilepsy by Pominova~\cite{pominova2018voxelwise} used the structural MRI and fMRI separately to classify the subjects with epilepsy. This work confirmed the feasibility of CNN in an end-to-end fashion which directly inputs the high dimensional neuroimaging data without feature engineering. Later, Yan et al. presented a \re{3D} CNN on the raw MRI image data to predict the benign epilepsy with centrotemporal spikes and achieved 89.80\% accuracy~\cite{Yan2018ADL}. Jiang et al. fused the features extracted from 2D and 3D PET ROIs respectively by three natural images pretrained networks (ResNet - 50, VGGNet - 16, Inception - V3), and a pulmonary nodules CT pretrained network (SVGG - C3D). Then, a fully connected layer was applied as a binary classification~\cite{jiang2019transfer}. Recently, deep learning method was applied to PET for the first time. Zhang et al. developed a Siamese CNN which is able to track the metabolic symmetricity of PET because the epileptic focus is strongly correlated to the high-dimensional inter-hemispheric symmetricity changes. Their proposed model had a high detection accuracy of 90\%~\cite{zhang2021deep}.

Besides the end-to-end supervised models, the unsupervised machine learning approaches have been used to automatically extract the latent representation of brain images, which can be further input to a classifier. Alaverdyan applied the unsupervised Siamese network to improve the latent feature extraction ~\cite{alaverdyan2018unsupervised, alaverdyan2020regularized}. Then the latent representation of the T1w MRI voxels inputs to a one-class SVM classifier for the outlier detection, which can determine whether it is the abnormal data. 
They \re{further} compared the \re{variants} of siamese networks, the stacked convolutional autoencoder, and the Wasserstein autoencoder, and the results suggested that the regularized siamese network has great potential to extract the hidden features and eventually achieved the best classification performance. Surprisingly, this method has even been shown to outperform human experts for MRI-negative cases.

\begin{table}[]
\caption{Classification of \re{epileptic} patients and \re{healthy} controls}
\vspace{-5mm}
\begin{center}
\small
\begin{tabular}{m{8em}m{10em}m{5em}m{10em}m{2em}m{2em}}
\\[-1.8ex]\hline 
\hline \\[-1.8ex] 
\re{\textbf{Task}} & \textbf{Features} & \textbf{Model} & \textbf{Performance} & \textbf{Data Size} & \textbf{Ref} \\
\hline
Classify generalized tonic-clonic seizures and the normal &Gray matter volume and fractional amplitude of low-frequency fluctuation differences on MRI &  SVM   & ACC of GM: 74.4\%\newline{}ACC of fALFF:83.7\%  & 14\textbackslash{}30 & ~\cite{wang2018multimodal}\\
\hline
Classify TLE and the normal & Extracting white matter fibers from DTI, generating weighted structural connectivity graphs & Logistic regression & ACC: 91.0\% & 17\textbackslash{}17 & ~\cite{ghazi2016structural}\\
\hline
Classify TLE and the normal &FreeSurfer extracted 936 features per subject from T1, T2 and DTI (624 mean intensity features and 312 intensity difference features) &  PCA+SVM & ACC : 88.9\% & 17\textbackslash{}19 & ~\cite{cantor2015detection}\\
\hline
Classify epilepsy and the normal & Curve (Perimeter of a VOI); VOI extracted from two lateral ventricles & K-Nearest
Neighbour & ACC : 78.5\% & 105\textbackslash{}105 & ~\cite{udomchaiporn20163}\\
\hline
Classify TLE and MCI & Spherical harmonics features from hippocampus & SVM & ACC : 84.2\% & 17\textbackslash{}20 & ~\cite{liedlgruber2019can}\\
\hline
Prediction of cognitive decline in TLE & Volumetry, local binary patterns, and wavelets & SVM   & ACC : 86.0\% & 9\textbackslash{}18 & ~\cite{holler2020prediction}\\
\hline 
\hline \\[-1.8ex] 

\re{\textbf{Task}} & \textbf{Inputs} & \textbf{Model} & \textbf{Performance} & \textbf{Data Size} & \textbf{Ref} \\
\hline
Classify epilepsy and the normal
 & MRI voxel & 3D-CNN & AUC: 61.0\% -76.0\%  & 21\textbackslash{}23 & ~\cite{pominova2018voxelwise}\\
\hline
Classify epilepsy and the normal & MRI pixel &  AE extract features + oc-SVM & Sensitivity: 60.0\%
 & 21\textbackslash{}75 & ~\cite{alaverdyan2018unsupervised}\\
\hline
\end{tabular}%
\centering
\label{tab:tasks-classification}
\end{center}
\end{table}

\subsection{Epilepsy Prognosis}

Computer-aided prognosis tasks \re{are important to guide} clinicians in appropriate treatment.
The prognosis of epilepsy is usually quantified with some clinical indexes, such as epidemiology-based mortality score in status epilepticus and Engel classification.
Machine learning models \re{can be trained to predict the treatment outcome based on the brain images from patients}, offering useful information for treatment decisions in advance.
Prognosis prediction usually refers to the classification task of classifying the postoperative state (\re{i.e}, no seizures or persistent seizures)
We list some work relevant to epilepsy prognosis in \textbf{Table~\ref{tab:tasks-prognosis}}.

\re{Many studies applied conventional models (that is, first extract features and then conduct classification)} to predict the outcome of surgery~\cite{memarian2015multimodal,bernhardt2015magnetic}.
For instance, Memarian et al. extracted 88 features (including demographic, clinical, electrophysiological, and structural MRI features) from mesial TLE patients, and \re{then input them into a SVM classifier, resulting} 95\% accuracy of prognostic prediction~\cite{memarian2015multimodal}. Bernhardt et al. applied K-means clustering into 4 subgroups based on the manually segmented mesiotemporal structures (e.g., hippocampus, amygdala, and entorhinal cortex) from T1w MRI~\cite{bernhardt2015magnetic}, \re{and then input} the surface morphology into a supervised LDA classifier, aiming at predicting the postsurgical outcome. The mean accuracy reached 92\% \re{in} the 4-class classification.

Apart from the \re{conventional} machine learning methods, end-to-end DNNs have also been applied to predict the most likely outcome of the treatment~\cite{gleichgerrcht2018deep,samson2018deep,munsell2015evaluation}. For example, \re{ Gleichgerrcht et al. trained a neural network with the the whole-brain connectome matrix obtained from DTI and the corresponding binary label (\textit{i.e.} surgical success or failure), achieving 88.0\% prediction accuracy for the surgical success cases and 79.0\% for the surgery failure cases~\cite{gleichgerrcht2018deep}.}

\begin{table}[]
\centering
\caption{\re{Prediction of prognosis in patients with epilepsy}}
\small
\begin{tabular}{m{8em}m{10em}m{5em}m{10em}m{2em}m{2em}}
\\[-1.8ex]\hline 
\hline \\[-1.8ex] 
\re{\textbf{Task}} & \textbf{Inputs} &  \textbf{Model} & \textbf{Performance} & \textbf{Data Size} & \textbf{Ref} \\
\hline
Predict surgical outcome & Structural MRI, sEEG, clinical as well as demographical features &  SVM   & ACC : 95.0\% & 20 & ~\cite{memarian2015multimodal}\\
\hline
Predict surgical outcome & Whole‐brain connectome from DTI & DNN & Surgical success prediction ratio: 88.0\%\newline{}Surgical failure prediction ratio: 79.0\% & 50 & ~\cite{gleichgerrcht2018deep}\\
\hline
Predict surgical outcome & Strength of each connection between all possible brain regions & DNN & ACC: 95.0\% & 50 & ~\cite{samson2018deep}\\
\hline
Predict surgical outcome & Surface morphology of hippocampus, amygdala, and entorhinal cortex & K-means clustering   & ACC: 92.0\% & 114 & ~\cite{bernhardt2015magnetic}\\
\hline
\end{tabular}%
\label{tab:tasks-prognosis}%
\end{table}%



\section{Discussion}
%
In short, among the various machine learning models, the feature engineering based \re{conventional machine learning} models account for a large proportion of applications, and the data-driven \re{deep learning} models for neuroimaging analysis have just emerged. Although the first study using deep learning models for epilepsy neuroimaging data \re{did not appear until 2018}~\cite{pominova2018voxelwise}, deep learning approach has been a success in computer vision, recommendation systems, and natural language processing~\cite{grys2017machine, voulodimos2018deep}. we can \re{imagine} that deep learning approach would play a more important role in epilepsy neuroimaging in the future. Along this line, neuroimaging will contribute to the computer-aided diagnosis \re{and prognosis} of epilepsy. 



However, there are obstacles when applying deep learning methods to the epilepsy neuroimaging field.
The challenges of deep learning in epilepsy neuroimaging have multiple manifolds. 
First, labeling data is challenging. The ambiguity in diagnosing particular epilepsy can \re{make it difficult} to label the disease type. \re{It is particularly difficult to mark epileptic lesions} unless the neurologist conducts longitudinal observations and confirms seizure-free after surgery. Moreover, patients with epilepsy may suffer from multiple complications, making it difficult to use one-hot vector labels to train machine learning models. Since the loss function defined by the discrepancy between network output and label is the key to the supervised learning algorithm based on gradient descent, noisy labels will inevitably bring serious negative effects. The consequences of mislabeling data can be fatal.
Second, The imbalance between classes is another challenge in epilepsy medical images. The fact that different types of epilepsy have different frequencies of occurrence leads to an imbalance between the types of epilepsy. For example, temporal lobe epilepsy is common, while focal epilepsy on the motor cortex is relatively rare~\cite{tellez2012review}. Therefore, the samples in the diagnosis task are tilted towards TLE, which might largely bias classification results.
Third, neuroimaging requires high spatial resolution to detect subtle abnormalities, and neuroimaging also needs a large field of view for locating abnormal brain areas of all spans. These two are difficult to meet in a single brain imaging modality. As a result, there are more and more modalities of epilepsy images\re{, which increases the difficulty of modeling.}
Finally, due to the diversity of neuroimaging equipment and patient conditions, medical images are usually heterogeneous; therefore,\re{it is very challenging to improve the generalization ability of machine learning models}. 
In fact, comprehensive validations of the \re{universality} and reliability of machine learning algorithms come at a cost. \re{When encountering MRI negative images, clinicians rarely have fully validated methods, so the algorithm is almost unreliable in clinical applications.} This distrust is further amplified by the shortcomings of machine learning \re{in the absence of} a clear physical explanation. When algorithm developers boast their high accuracy rate, it is still difficult for clinicians to understand why it \re{succeeded, and those few failed samples did not explain why or when they failed.}

The cutting-edge machine learning models provide potential tools for addressing the above challenges. For example, generative adversarial networks (GAN)~\cite{Goodfellow2014GenerativeAN,Mirza2014ConditionalGA,Radford2016UnsupervisedRL} might be beneficial for data augmentation to address the problem of unbalanced neuroimaging samples~\cite{Ma2020CombiningDW}. Likelihood-based generative models (\textit{e.g.} VAEs~\cite{Kingma2014AutoEncodingVB,Rezende2014StochasticBA},  Pixel CNN~\cite{Oord2016ConditionalIG} and Glow~\cite{Kingma2018GlowGF}) have been reported to be highly robust to the anomaly data and out-of-distribution (OOD) inputs and therefore can be used to detect the anomaly samples. Anomaly detection has wide medical applications. For instance, when an untypical disease is not seen in the training samples, likelihood-based generative models equipped with medical OOD detection might deal with noisy labels and largely improve the robustness~\cite{Ran2020BigeminalPV,Ran2020DetectingOS,Cohen2020ABO,Ren2019LikelihoodRF}.
The recent advancements in graph convolutional networks (GCN) can work with the non-regular data structures. In contrast to the node classification tasks of medical image analysis~\cite{zhou2020graph, wu2020comprehensive}, GCN can handle unstructured data~\cite{kipf2016semi}. As a generalization of CNN, GCN is compatible with \re{neuroimaging} data in the non-Euclidean domain \re{such as the BOLD signal and the DWI streamline between different brain areas}. Segmentation~\cite{gopinath2019graph}, localization~\cite{zeng2019graph}, and prediction~\cite{song2020graph} tasks have been tested \re{applicable} to GCN, although the performance of GCN in epilepsy data \re{has yet to be proven}. 


On the other hand, the challenges and special needs in epilepsy applications can greatly motivate the development of machine learning methods. For example, Zhao et al.~\cite{zhao2016non} came up with a new approach to solve the problems of patients' heterogeneity by combining a restricted Boltzmann machine with a Bayesian non-parametric hybrid model. The heterogeneity of patients with epilepsy requires personalized medical care, which calls for accurate diagnosis of the stratification of each patient, such as the cause of epilepsy, the location, and symptoms of epilepsy~\cite{alaverdyan2018feature, zhao2017addressing}. Since the prognosis of epilepsy is vital for the patient's quality of life, the longitudinal medical records of each patient should be recorded and analyzed. These are strong drives of machine learning theory and method. 

\section{Conclusion}
Just as machine learning, especially deep learning, brings benefits to machine vision and natural language processing, machine learning on neuroimaging promises to extend those same benefits to epilepsy diagnosis and prognosis prediction. However, clinic-oriented machine learning applications have their unique context, such as the small data size, the low certainty in labeling, and the large heterogeneity across patients. It requires more efforts from multi-disciplinary experts on learning theory, neuroimaging, and epilepsy clinical applications.

\section*{Acknowledgement}
This work was funded in part by the National Natural Science Foundation of China (62001205), Guangdong Natural Science Foundation Joint Fund (2019A1515111038), Shenzhen Science and Technology Innovation Committee (20200925155957004, KCXFZ2020122117340001, SGDX2020110309280100), Shenzhen Key Laboratory of Smart Healthcare Engineering (ZDSYS20200811144003009).\\

\noindent The authors declare that they have no competing interests.

\setlength{\baselineskip}{12pt}

\begin{thebibliography}{100}

\bibitem{fisher2014ilae}
Robert~S Fisher, Carlos Acevedo, Alexis Arzimanoglou, Alicia Bogacz, J~Helen
  Cross, Christian~E Elger, Jerome Engel~Jr, Lars Forsgren, Jacqueline~A
  French, Mike Glynn, et~al.
\newblock Ilae official report: a practical clinical definition of epilepsy.
\newblock {\em Epilepsia}, 55(4):475--482, 2014.

\bibitem{fiest2017prevalence}
Kirsten~M Fiest, Khara~M Sauro, Samuel Wiebe, Scott~B Patten, Churl-Su Kwon,
  Jonathan Dykeman, Tamara Pringsheim, Diane~L Lorenzetti, and Nathalie
  Jett{\'e}.
\newblock Prevalence and incidence of epilepsy: a systematic review and
  meta-analysis of international studies.
\newblock {\em Neurology}, 88(3):296--303, 2017.

\bibitem{eadie2012shortcomings}
Mervyn~J Eadie.
\newblock Shortcomings in the current treatment of epilepsy.
\newblock {\em Expert review of neurotherapeutics}, 12(12):1419--1427, 2012.

\bibitem{schramm2008surgery}
Johannes Schramm and Hans Clusmann.
\newblock The surgery of epilepsy.
\newblock {\em Neurosurgery}, 62(suppl\_2):SHC--463, 2008.

\bibitem{spencer1989corpus}
SS~SPENCER.
\newblock Corpus callosotomy in the treatment of medically intractable
  secondarily generalized seizures of children.
\newblock {\em Cleve Clin J Med}, 56(1):S69--S78, 1989.

\bibitem{clusmann2002prognostic}
Hans Clusmann, Johannes Schramm, Thomas Kral, Christoph Helmstaedter, Burkhard
  Ostertun, Rolf Fimmers, Dorothee Haun, and Christian~E Elger.
\newblock Prognostic factors and outcome after different types of resection for
  temporal lobe epilepsy.
\newblock {\em Journal of neurosurgery}, 97(5):1131--1141, 2002.

\bibitem{juhasz2020utility}
Csaba Juh{\'a}sz and Fl{\'o}ra John.
\newblock Utility of mri, pet, and ictal spect in presurgical evaluation of
  non-lesional pediatric epilepsy.
\newblock {\em Seizure}, 77:15--28, 2020.

\bibitem{zijlmans2019changing}
Maeike Zijlmans, Willemiek Zweiphenning, and Nicole van Klink.
\newblock Changing concepts in presurgical assessment for epilepsy surgery.
\newblock {\em Nature Reviews Neurology}, 15(10):594--606, 2019.

\bibitem{nunes2012diagnosis}
Vanessa~Delgado Nunes, Laura Sawyer, Julie Neilson, Grammati Sarri, and J~Helen
  Cross.
\newblock Diagnosis and management of the epilepsies in adults and children:
  summary of updated nice guidance.
\newblock {\em Bmj}, 344:e281, 2012.

\bibitem{west2019surgery}
Siobhan West, Sarah~J Nevitt, Jennifer Cotton, Sacha Gandhi, Jennifer Weston,
  Ajay Sudan, Roberto Ramirez, and Richard Newton.
\newblock Surgery for epilepsy.
\newblock {\em Cochrane Database of Systematic Reviews}, (6), 2019.

\bibitem{zhao2017role}
Xu~Zhao, Zhiqiang Zhou, Wenzhen Zhu, and Hongbing Xiang.
\newblock Role of conventional magnetic resonance imaging in the screening of
  epilepsy with structural abnormalities: a pictorial essay.
\newblock {\em American Journal of Nuclear Medicine and Molecular Imaging},
  7(3):126, 2017.

\bibitem{la2009pet}
C~La~Foug{\`e}re, A~Rominger, S~F{\"o}rster, J~Geisler, and P~Bartenstein.
\newblock Pet and spect in epilepsy: a critical review.
\newblock {\em Epilepsy \& Behavior}, 15(1):50--55, 2009.

\bibitem{tellez2010surgical}
Jos{\'e}~F T{\'e}llez-Zenteno, Lizbeth~Hern{\'a}ndez Ronquillo, Farzad
  Moien-Afshari, and Samuel Wiebe.
\newblock Surgical outcomes in lesional and non-lesional epilepsy: a systematic
  review and meta-analysis.
\newblock {\em Epilepsy research}, 89(2-3):310--318, 2010.

\bibitem{shoeibi2021applications}
Afshin Shoeibi, Navid Ghassemi, Marjane Khodatars, Mahboobeh Jafari, Parisa
  Moridian, Roohallah Alizadehsani, Ali Khadem, Yinan Kong, Assef Zare,
  Juan~Manuel Gorriz, et~al.
\newblock Applications of epileptic seizures detection in neuroimaging
  modalities using deep learning techniques: Methods, challenges, and future
  works.
\newblock {\em arXiv preprint arXiv:2105.14278}, 2021.

\bibitem{keller2008voxel}
Simon~Sean Keller and Neil Roberts.
\newblock Voxel-based morphometry of temporal lobe epilepsy: an introduction
  and review of the literature.
\newblock {\em Epilepsia}, 49(5):741--757, 2008.

\bibitem{acharya2019characterization}
U~Rajendra Acharya, Yuki Hagiwara, Sunny~Nitin Deshpande, S~Suren, Joel En~Wei
  Koh, Shu~Lih Oh, N~Arunkumar, Edward~J Ciaccio, and Choo~Min Lim.
\newblock Characterization of focal eeg signals: a review.
\newblock {\em Future Generation Computer Systems}, 91:290--299, 2019.

\bibitem{boonyakitanont2020review}
Poomipat Boonyakitanont, Apiwat Lek-Uthai, Krisnachai Chomtho, and Jitkomut
  Songsiri.
\newblock A review of feature extraction and performance evaluation in
  epileptic seizure detection using eeg.
\newblock {\em Biomedical Signal Processing and Control}, 57:101702, 2020.

\bibitem{shoeibi2021epileptic}
Afshin Shoeibi, Marjane Khodatars, Navid Ghassemi, Mahboobeh Jafari, Parisa
  Moridian, Roohallah Alizadehsani, Maryam Panahiazar, Fahime Khozeimeh, Assef
  Zare, Hossein Hosseini-Nejad, et~al.
\newblock Epileptic seizures detection using deep learning techniques: A
  review.
\newblock {\em International Journal of Environmental Research and Public
  Health}, 18(11):5780, 2021.

\bibitem{abbasi2019machine}
Bardia Abbasi and Daniel~M Goldenholz.
\newblock Machine learning applications in epilepsy.
\newblock {\em Epilepsia}, 60(10):2037--2047, 2019.

\bibitem{kini2016computational}
Lohith~G Kini, James~C Gee, and Brian Litt.
\newblock Computational analysis in epilepsy neuroimaging: a survey of features
  and methods.
\newblock {\em NeuroImage: Clinical}, 11:515--529, 2016.

\bibitem{giedd2004structural}
Jay~N Giedd.
\newblock Structural magnetic resonance imaging of the adolescent brain.
\newblock {\em Annals of the new york academy of sciences}, 1021(1):77--85,
  2004.

\bibitem{adler2017novel}
Sophie Adler, Konrad Wagstyl, Roxana Gunny, Lisa Ronan, David Carmichael,
  J~Helen Cross, Paul~C Fletcher, and Torsten Baldeweg.
\newblock Novel surface features for automated detection of focal cortical
  dysplasias in paediatric epilepsy.
\newblock {\em NeuroImage: Clinical}, 14:18--27, 2017.

\bibitem{wagner2011morphometric}
Jan Wagner, Bernd Weber, Horst Urbach, Christian~E Elger, and Hans-J{\"u}rgen
  Huppertz.
\newblock Morphometric mri analysis improves detection of focal cortical
  dysplasia type ii.
\newblock {\em Brain}, 134(10):2844--2854, 2011.

\bibitem{colombo2009imaging}
Nadia Colombo, Noriko Salamon, Charles Raybaud, {\c{C}}igdem {\"O}zkara, and
  A~James Barkovich.
\newblock Imaging of malformations of cortical development.
\newblock {\em Epileptic Disorders}, 11(3):194--205, 2009.

\bibitem{steven2014diffusion}
Andrew~J Steven, Jiachen Zhuo, and Elias~R Melhem.
\newblock Diffusion kurtosis imaging: an emerging technique for evaluating the
  microstructural environment of the brain.
\newblock {\em American journal of roentgenology}, 202(1):W26--W33, 2014.

\bibitem{arab2018principles}
Anas Arab, Anna Wojna-Pelczar, Amit Khairnar, Nikoletta Szab{\'o}, and Jana
  Ruda-Kucerova.
\newblock Principles of diffusion kurtosis imaging and its role in early
  diagnosis of neurodegenerative disorders.
\newblock {\em Brain research bulletin}, 139:91--98, 2018.

\bibitem{kwong1992dynamic}
Kenneth~K Kwong, John~W Belliveau, David~A Chesler, Inna~E Goldberg, Robert~M
  Weisskoff, Brigitte~P Poncelet, David~N Kennedy, Bernice~E Hoppel, Mark~S
  Cohen, and Robert Turner.
\newblock Dynamic magnetic resonance imaging of human brain activity during
  primary sensory stimulation.
\newblock {\em Proceedings of the National Academy of Sciences},
  89(12):5675--5679, 1992.

\bibitem{sweet1955localization}
William~H Sweet and Gordon~L Brownell.
\newblock Localization of intracranial lesions by scanning with
  positron-emitting arsenic.
\newblock {\em Journal of the American Medical Association},
  157(14):1183--1188, 1955.

\bibitem{velez2012neuroimaging}
Naymee~J Velez-Ruiz and Joshua~P Klein.
\newblock Neuroimaging in the evaluation of epilepsy.
\newblock In {\em Seminars in neurology}, volume~32, pages 361--373. Thieme
  Medical Publishers, 2012.

\bibitem{bishop2006pattern}
Christopher~M Bishop.
\newblock {\em Pattern recognition and machine learning}.
\newblock springer, 2006.

\bibitem{wang2018multimodal}
Jianping Wang, Yongxin Li, Ya~Wang, and Wenhua Huang.
\newblock Multimodal data and machine learning for detecting specific
  biomarkers in pediatric epilepsy patients with generalized tonic-clonic
  seizures.
\newblock {\em Frontiers in neurology}, 9:1038, 2018.

\bibitem{cantor2015detection}
Diego Cantor-Rivera, Ali~R Khan, Maged Goubran, Seyed~M Mirsattari, and Terry~M
  Peters.
\newblock Detection of temporal lobe epilepsy using support vector machines in
  multi-parametric quantitative mr imaging.
\newblock {\em Computerized Medical Imaging and Graphics}, 41:14--28, 2015.

\bibitem{alaverdyan2020regularized}
Zaruhi Alaverdyan, Julien Jung, Romain Bouet, and Carole Lartizien.
\newblock Regularized siamese neural network for unsupervised outlier detection
  on brain multiparametric magnetic resonance imaging: application to epilepsy
  lesion screening.
\newblock {\em Medical Image Analysis}, 60:101618, 2020.

\bibitem{farazi2017lateralization}
Mohammad Farazi and Hamid Soltanian-Zadeh.
\newblock Lateralization and prognosis of temporal lobe epilepsy patients by
  shape analysis of hippocampus via signed poisson mapping.
\newblock In {\em 2017 10th Iranian Conference on Machine Vision and Image
  Processing (MVIP)}, pages 203--208. IEEE, 2017.

\bibitem{memarian2015multimodal}
Negar Memarian, Sally Kim, Sandra Dewar, Jerome Engel~Jr, and Richard~J Staba.
\newblock Multimodal data and machine learning for surgery outcome prediction
  in complicated cases of mesial temporal lobe epilepsy.
\newblock {\em Computers in biology and medicine}, 64:67--78, 2015.

\bibitem{gleichgerrcht2018deep}
Ezequiel Gleichgerrcht, Brent Munsell, Sonal Bhatia, William~A Vandergrift~III,
  Chris Rorden, Carrie McDonald, Jonathan Edwards, Ruben Kuzniecky, and
  Leonardo Bonilha.
\newblock Deep learning applied to whole-brain connectome to determine seizure
  control after epilepsy surgery.
\newblock {\em Epilepsia}, 59(9):1643--1654, 2018.

\bibitem{munsell2019relationship}
BC~Munsell, G~Wu, J~Fridriksson, K~Thayer, N~Mofrad, N~Desisto, Dinggang Shen,
  and L~Bonilha.
\newblock Relationship between neuronal network architecture and naming
  performance in temporal lobe epilepsy: A connectome based approach using
  machine learning.
\newblock {\em Brain and Language}, 193:45--57, 2019.

\bibitem{sayed2021characterization}
Tasneem Abdulrazig~Mohamed Sayed, Fatima~Yousif Mohammed, and Maha~Esmeal
  Ahmed.
\newblock Characterization of hippocampus on epileptic patients on mri using
  texture analysis techniques.
\newblock {\em International Journal of Research-GRANTHAALAYAH}, 9(1):164--168,
  2021.

\bibitem{wagstyl2020planning}
Konrad Wagstyl, Sophie Adler, Birgit Pimpel, Aswin Chari, Kiran Seunarine, Sara
  Lorio, Rachel Thornton, Torsten Baldeweg, and Martin Tisdall.
\newblock Planning stereoelectroencephalography using automated lesion
  detection: Retrospective feasibility study.
\newblock {\em Epilepsia}, 61(7):1406--1416, 2020.

\bibitem{kim2014multivariate}
Hosung Kim, Boris~C Bernhardt, Jessie Kulaga-Yoskovitz, Benoit Caldairou,
  Andrea Bernasconi, and Neda Bernasconi.
\newblock Multivariate hippocampal subfield analysis of local mri intensity and
  volume: application to temporal lobe epilepsy.
\newblock In {\em International Conference on Medical Image Computing and
  Computer-Assisted Intervention}, pages 170--178. Springer, 2014.

\bibitem{sahebzamani2019machine}
Ghazal Sahebzamani, Mansour Saffar, and Hamid Soltanian-Zadeh.
\newblock Machine learning based analysis of structural mri for epilepsy
  diagnosis.
\newblock In {\em 2019 4th International Conference on Pattern Recognition and
  Image Analysis (IPRIA)}, pages 58--63. IEEE, 2019.

\bibitem{jafari2010flair}
Kourosh Jafari-Khouzani, Kost Elisevich, Suresh Patel, Brien Smith, and Hamid
  Soltanian-Zadeh.
\newblock Flair signal and texture analysis for lateralizing mesial temporal
  lobe epilepsy.
\newblock {\em Neuroimage}, 49(2):1559--1571, 2010.

\bibitem{del2017using}
John Del~Gaizo, Neda Mofrad, Jens~H Jensen, David Clark, Russell Glenn, Joseph
  Helpern, and Leonardo Bonilha.
\newblock Using machine learning to classify temporal lobe epilepsy based on
  diffusion mri.
\newblock {\em Brain and behavior}, 7(10):e00801, 2017.

\bibitem{huang2020identifying}
Jianjun Huang, Jiahui Xu, Li~Kang, and Tijiang Zhang.
\newblock Identifying epilepsy based on deep learning using dki images.
\newblock {\em Frontiers in Human Neuroscience}, 14:465, 2020.

\bibitem{torlay2017machine}
L~Torlay, Marcela Perrone-Bertolotti, Elizabeth Thomas, and Monica Baciu.
\newblock Machine learning--xgboost analysis of language networks to classify
  patients with epilepsy.
\newblock {\em Brain informatics}, 4(3):159--169, 2017.

\bibitem{jiang2019transfer}
Huiyan Jiang, Feifei Gao, Xiaoyu Duan, Zhiqi Bai, Zhiguo Wang, Xiaoqi Ma, and
  Yen-Wei Chen.
\newblock Transfer learning and fusion model for classification of epileptic
  pet images.
\newblock In {\em Innovation in Medicine and Healthcare Systems, and
  Multimedia}, pages 71--79. Springer, 2019.

\bibitem{motoda2002feature}
Hiroshi Motoda and Huan Liu.
\newblock Feature selection, extraction and construction.
\newblock {\em Communication of IICM (Institute of Information and Computing
  Machinery, Taiwan) Vol}, 5(67-72):2, 2002.

\bibitem{https://doi.org/10.1111/j.1469-1809.1936.tb02137.x}
R.~A. FISHER.
\newblock The use of multiple measurements in taxonomic problems.
\newblock {\em Annals of Eugenics}, 7(2):179--188, 1936.

\bibitem{McLachlan1992DiscriminantAA}
G.~McLachlan.
\newblock Discriminant analysis and statistical pattern recognition.
\newblock 1992.

\bibitem{10.2307/2290989}
Trevor Hastie, Robert Tibshirani, and Andreas Buja.
\newblock Flexible discriminant analysis by optimal scoring.
\newblock {\em Journal of the American Statistical Association},
  89(428):1255--1270, 1994.

\bibitem{friedman1989regularized}
Jerome~H Friedman.
\newblock Regularized discriminant analysis.
\newblock {\em Journal of the American Statistical Association},
  84(405):165--175, 1989.

\bibitem{10.1023/A:1010933404324}
Leo Breiman.
\newblock Random forests.
\newblock {\em Mach. Learn.}, 45(1):5–32, October 2001.

\bibitem{breiman1996bagging}
Leo Breiman.
\newblock Bagging predictors.
\newblock {\em Machine Learning}, 24(2):123--140, August 1996.

\bibitem{cortes1995support}
Corinna Cortes and Vladimir Vapnik.
\newblock Support-vector networks.
\newblock {\em Machine learning}, 20(3):273--297, 1995.

\bibitem{Smola2004ATO}
Alex Smola and B.~Sch{\"o}lkopf.
\newblock A tutorial on support vector regression.
\newblock {\em Statistics and Computing}, 14:199--222, 2004.

\bibitem{10.5555/3009657.3009740}
Bernhard Sch\"{o}lkopf, Robert Williamson, Alex Smola, John Shawe-Taylor, and
  John Platt.
\newblock Support vector method for novelty detection.
\newblock In {\em Proceedings of the 12th International Conference on Neural
  Information Processing Systems}, NIPS'99, page 582–588, Cambridge, MA, USA,
  1999. MIT Press.

\bibitem{Suykens2004LeastSS}
J.~Suykens and J.~Vandewalle.
\newblock Least squares support vector machine classifiers.
\newblock {\em Neural Processing Letters}, 9:293--300, 2004.

\bibitem{Suykens2002WeightedLS}
J.~Suykens, J.~D. Brabanter, L.~Lukas, and J.~Vandewalle.
\newblock Weighted least squares support vector machines: robustness and sparse
  approximation.
\newblock {\em Neurocomputing}, 48:85--105, 2002.

\bibitem{Liu2016RampLL}
D.~Liu, Y.~Shi, Y.~Tian, and X.~Huang.
\newblock Ramp loss least squares support vector machine.
\newblock {\em J. Comput. Sci.}, 14:61--68, 2016.

\bibitem{Lin2002FuzzySV}
C.~Lin and Sheng-De Wang.
\newblock Fuzzy support vector machines.
\newblock {\em IEEE transactions on neural networks}, 13 2:464--71, 2002.

\bibitem{Lin2005FuzzySV}
C.~fu~Lin and S.~de~Wang.
\newblock Fuzzy support vector machines with automatic membership setting.
\newblock 2005.

\bibitem{Lin2004TrainingAF}
C.~Lin and Sheng-De Wang.
\newblock Training algorithms for fuzzy support vector machines with noisy
  data.
\newblock {\em Pattern Recognit. Lett.}, 25:1647--1656, 2004.

\bibitem{Yang2005WeightedSV}
XuLei Yang, Q.~Song, and Yue Wang.
\newblock Weighted support vector machine for data classification.
\newblock {\em Proceedings. 2005 IEEE International Joint Conference on Neural
  Networks, 2005.}, 2:859--864 vol. 2, 2005.

\bibitem{Cevikalp2017LargescaleRT}
Hakan Cevikalp and Vojtech Franc.
\newblock Large-scale robust transductive support vector machines.
\newblock {\em Neurocomputing}, 235:199--209, 2017.

\bibitem{Jayadeva2007TwinSV}
Jayadeva, Reshma Khemchandani, and S.~Chandra.
\newblock Twin support vector machines for pattern classification.
\newblock {\em IEEE Transactions on Pattern Analysis and Machine Intelligence},
  29:905--910, 2007.

\bibitem{2020An}
Chenyang Shen, Dan Nguyen, Zhiguo Zhou, Steve~B Jiang, and Xun Jia.
\newblock An introduction to deep learning in medical physics: advantages,
  potential, and challenges.
\newblock {\em Physics in Medicine and Biology}, 65(5):05TR01--, 2020.

\bibitem{hring2020expressivity}
Ingo Gühring, Mones Raslan, and Gitta Kutyniok.
\newblock Expressivity of deep neural networks, 2020.

\bibitem{Rumelhart1986LearningRB}
D.~Rumelhart, Geoffrey~E. Hinton, and R.~J. Williams.
\newblock Learning representations by back-propagating errors.
\newblock {\em Nature}, 323:533--536, 1986.

\bibitem{billings1992properties}
SA~Billings, HB~Jamaluddin, and S~Chen.
\newblock Properties of neural networks with applications to modelling
  non-linear dynamical systems.
\newblock {\em International Journal of Control}, 55(1):193--224, 1992.

\bibitem{Glasmachers2017LimitsOE}
T.~Glasmachers.
\newblock Limits of end-to-end learning.
\newblock In {\em ACML}, 2017.

\bibitem{guo2016deep}
Yanming Guo, Yu~Liu, Ard Oerlemans, Songyang Lao, Song Wu, and Michael~S Lew.
\newblock Deep learning for visual understanding: A review.
\newblock {\em Neurocomputing}, 187:27--48, 2016.

\bibitem{mohri2018foundations}
Mehryar Mohri, Afshin Rostamizadeh, and Ameet Talwalkar.
\newblock {\em Foundations of machine learning}.
\newblock MIT press, 2018.

\bibitem{hinton1999unsupervised}
Geoffrey~E Hinton, Terrence~Joseph Sejnowski, et~al.
\newblock {\em Unsupervised learning: foundations of neural computation}.
\newblock MIT press, 1999.

\bibitem{LeCun1998GradientbasedLA}
Y.~LeCun, L.~Bottou, Yoshua Bengio, and P.~Haffner.
\newblock Gradient-based learning applied to document recognition.
\newblock 1998.

\bibitem{Krizhevsky2017ImageNetCW}
A.~Krizhevsky, Ilya Sutskever, and Geoffrey~E. Hinton.
\newblock Imagenet classification with deep convolutional neural networks.
\newblock In {\em CACM}, 2017.

\bibitem{pominova2018voxelwise}
Marina Pominova, Alexey Artemov, Maksim Sharaev, Ekaterina Kondrateva,
  Alexander Bernstein, and Evgeny Burnaev.
\newblock Voxelwise 3d convolutional and recurrent neural networks for epilepsy
  and depression diagnostics from structural and functional mri data.
\newblock In {\em 2018 IEEE International Conference on Data Mining Workshops
  (ICDMW)}, pages 299--307. IEEE, 2018.

\bibitem{kumar2018u}
Pulkit Kumar, Pravin Nagar, Chetan Arora, and Anubha Gupta.
\newblock U-segnet: fully convolutional neural network based automated brain
  tissue segmentation tool.
\newblock In {\em 2018 25th IEEE International Conference on Image Processing
  (ICIP)}, pages 3503--3507. IEEE, 2018.

\bibitem{carmo2021hippocampus}
Diedre Carmo, Bruna Silva, Clarissa Yasuda, Let{\'\i}cia Rittner, Roberto
  Lotufo, Alzheimer's Disease~Neuroimaging Initiative, et~al.
\newblock Hippocampus segmentation on epilepsy and alzheimer's disease studies
  with multiple convolutional neural networks.
\newblock {\em Heliyon}, 7(2):e06226, 2021.

\bibitem{Hahnloser2000DigitalSA}
R.~Hahnloser, R.~Sarpeshkar, M.~Mahowald, R.~Douglas, and H.~S. Seung.
\newblock Digital selection and analogue amplification coexist in a
  cortex-inspired silicon circuit.
\newblock {\em Nature}, 405:947--951, 2000.

\bibitem{Nair2010RectifiedLU}
V.~Nair and Geoffrey~E. Hinton.
\newblock Rectified linear units improve restricted boltzmann machines.
\newblock In {\em ICML}, 2010.

\bibitem{Srivastava2014DropoutAS}
Nitish Srivastava, Geoffrey~E. Hinton, A.~Krizhevsky, Ilya Sutskever, and
  R.~Salakhutdinov.
\newblock Dropout: a simple way to prevent neural networks from overfitting.
\newblock {\em J. Mach. Learn. Res.}, 15:1929--1958, 2014.

\bibitem{Ioffe2015BatchNA}
S.~Ioffe and Christian Szegedy.
\newblock Batch normalization: Accelerating deep network training by reducing
  internal covariate shift.
\newblock In {\em ICML}, 2015.

\bibitem{PICCIALLI2021111}
Francesco Piccialli, Vittorio~Di Somma, Fabio Giampaolo, Salvatore Cuomo, and
  Giancarlo Fortino.
\newblock A survey on deep learning in medicine: Why, how and when?
\newblock {\em Information Fusion}, 66:111 -- 137, 2021.

\bibitem{Vincent2008ExtractingAC}
Pascal Vincent, H.~Larochelle, Yoshua Bengio, and Pierre-Antoine Manzagol.
\newblock Extracting and composing robust features with denoising autoencoders.
\newblock In {\em ICML '08}, 2008.

\bibitem{Ranzato2006EfficientLO}
Marc'Aurelio Ranzato, Christopher~S. Poultney, S.~Chopra, and Y.~LeCun.
\newblock Efficient learning of sparse representations with an energy-based
  model.
\newblock In {\em NIPS}, 2006.

\bibitem{Kingma2014AutoEncodingVB}
Diederik~P. Kingma and M.~Welling.
\newblock Auto-encoding variational bayes.
\newblock {\em International Conference on Learning Representations}, 2014.

\bibitem{Rezende2014StochasticBA}
Danilo~Jimenez Rezende, S.~Mohamed, and Daan Wierstra.
\newblock Stochastic backpropagation and approximate inference in deep
  generative models.
\newblock In {\em ICML}, 2014.

\bibitem{Makhzani2015AdversarialA}
Alireza Makhzani, Jonathon Shlens, Navdeep Jaitly, and Ian~J. Goodfellow.
\newblock Adversarial autoencoders.
\newblock {\em ArXiv}, abs/1511.05644, 2015.

\bibitem{Shin2013StackedAF}
Hoo-Chang Shin, M.~Orton, D.~Collins, S.~Doran, and M.~Leach.
\newblock Stacked autoencoders for unsupervised feature learning and multiple
  organ detection in a pilot study using 4d patient data.
\newblock {\em IEEE Transactions on Pattern Analysis and Machine Intelligence},
  35:1930--1943, 2013.

\bibitem{MLVisuals}
dair.ai.
\newblock Ml visuals.
\newblock \url{https://github.com/dair-ai/ml-visuals} Accessed January 15,
  2021.

\bibitem{ronneberger2015u}
Olaf Ronneberger, Philipp Fischer, and Thomas Brox.
\newblock U-net: Convolutional networks for biomedical image segmentation.
\newblock In {\em International Conference on Medical image computing and
  computer-assisted intervention}, pages 234--241. Springer, 2015.

\bibitem{Dev2019AutomaticDA}
K.~Dev, Pawan~S. Jogi, S.~Niyas, S.~Vinayagamani, C.~Kesavadas, and J.~Rajan.
\newblock Automatic detection and localization of focal cortical dysplasia
  lesions in mri using fully convolutional neural network.
\newblock {\em Biomed. Signal Process. Control.}, 52:218--225, 2019.

\bibitem{onofrey2018segmenting}
John~A Onofrey, Lawrence~H Staib, and Xenophon Papademetris.
\newblock Segmenting the brain surface from ct images with artifacts using
  locally oriented appearance and dictionary learning.
\newblock {\em IEEE transactions on medical imaging}, 38(2):596--607, 2018.

\bibitem{thom2014hippocampal}
Maria Thom.
\newblock Hippocampal sclerosis in epilepsy: a neuropathology review.
\newblock {\em Neuropathology and applied neurobiology}, 40(5):520--543, 2014.

\bibitem{taylor1971focal}
DC~Taylor, MA~Falconer, CJ~Bruton, and JAN Corsellis.
\newblock Focal dysplasia of the cerebral cortex in epilepsy.
\newblock {\em Journal of Neurology, Neurosurgery \& Psychiatry},
  34(4):369--387, 1971.

\bibitem{el2013computer}
Meriem El~Azami, Alexander Hammers, Nicolas Costes, and Carole Lartizien.
\newblock Computer aided diagnosis of intractable epilepsy with mri imaging
  based on textural information.
\newblock In {\em 2013 International Workshop on Pattern Recognition in
  Neuroimaging}, pages 90--93. IEEE, 2013.

\bibitem{el2016detection}
Meriem El~Azami, Alexander Hammers, Julien Jung, Nicolas Costes, Romain Bouet,
  and Carole Lartizien.
\newblock Detection of lesions underlying intractable epilepsy on t1-weighted
  mri as an outlier detection problem.
\newblock {\em PloS one}, 11(9):e0161498, 2016.

\bibitem{ahmed2014hierarchical}
Bilal Ahmed, Thomas Thesen, Karen Blackmon, Yijun Zhao, Orrin Devinsky, Ruben
  Kuzniecky, and Carla Brodley.
\newblock Hierarchical conditional random fields for outlier detection: an
  application to detecting epileptogenic cortical malformations.
\newblock In {\em International Conference on Machine Learning}, pages
  1080--1088, 2014.

\bibitem{jin2018automated}
Bo~Jin, Balu Krishnan, Sophie Adler, Konrad Wagstyl, Wenhan Hu, Stephen Jones,
  Imad Najm, Andreas Alexopoulos, Kai Zhang, Jianguo Zhang, et~al.
\newblock Automated detection of focal cortical dysplasia type ii with
  surface-based magnetic resonance imaging postprocessing and machine learning.
\newblock {\em Epilepsia}, 59(5):982--992, 2018.

\bibitem{gill2017automated}
Ravnoor~S Gill, Seok-Jun Hong, Fatemeh Fadaie, Benoit Caldairou, Boris
  Bernhardt, Neda Bernasconi, and Andrea Bernasconi.
\newblock Automated detection of epileptogenic cortical malformations using
  multimodal mri.
\newblock In {\em Deep Learning in Medical Image Analysis and Multimodal
  Learning for Clinical Decision Support}, pages 349--356. Springer, 2017.

\bibitem{gill2018deep}
Ravnoor~S Gill, Seok-Jun Hong, Fatemeh Fadaie, Benoit Caldairou, Boris~C
  Bernhardt, Carmen Barba, Armin Brandt, Vanessa~C Coelho, Ludovico
  d’Incerti, Matteo Lenge, et~al.
\newblock Deep convolutional networks for automated detection of epileptogenic
  brain malformations.
\newblock In {\em International Conference on Medical Image Computing and
  Computer-Assisted Intervention}, pages 490--497. Springer, 2018.

\bibitem{xu2019objective}
Haotian Xu, Ming Dong, Min-Hee Lee, Nolan O’Hara, Eishi Asano, and Jeong-Won
  Jeong.
\newblock Objective detection of eloquent axonal pathways to minimize
  postoperative deficits in pediatric epilepsy surgery using diffusion
  tractography and convolutional neural networks.
\newblock {\em IEEE Transactions on Medical Imaging}, 38(8):1910--1922, 2019.

\bibitem{lee2020novel}
Min-Hee Lee, Nolan O'Hara, Masaki Sonoda, Naoto Kuroda, Csaba Juhasz, Eishi
  Asano, Ming Dong, and Jeong-Won Jeong.
\newblock Novel deep learning network analysis of electrical stimulation
  mapping-driven diffusion mri tractography to improve preoperative evaluation
  of pediatric epilepsy.
\newblock {\em IEEE Transactions on Biomedical Engineering}, 2020.

\bibitem{roland2019comparison}
Jarod~L Roland, Carl~D Hacker, Abraham~Z Snyder, Joshua~S Shimony, John~M
  Zempel, David~D Limbrick, Matthew~D Smyth, and Eric~C Leuthardt.
\newblock A comparison of resting state functional magnetic resonance imaging
  to invasive electrocortical stimulation for sensorimotor mapping in pediatric
  patients.
\newblock {\em NeuroImage: Clinical}, 23:101850, 2019.

\bibitem{zhao2016non}
Yijun Zhao, Bilal Ahmed, Thomas Thesen, Karen~E Blackmon, Jennifer~G Dy,
  Carla~E Brodley, Ruben Kuzniekcy, and Orrin Devinsky.
\newblock A non-parametric approach to detect epileptogenic lesions using
  restricted boltzmann machines.
\newblock In {\em Proceedings of the 22nd ACM SIGKDD International Conference
  on Knowledge Discovery and Data Mining}, pages 373--382, 2016.

\bibitem{hosseini2014support}
Mohammad-Parsa Hosseini, Mohammad~R Nazem-Zadeh, Fariborz Mahmoudi, Hao Ying,
  and Hamid Soltanian-Zadeh.
\newblock Support vector machine with nonlinear-kernel optimization for
  lateralization of epileptogenic hippocampus in mr images.
\newblock In {\em 2014 36th Annual International Conference of the IEEE
  Engineering in Medicine and Biology Society}, pages 1047--1050. IEEE, 2014.

\bibitem{kamiya2015machine}
Kouhei Kamiya, Shiori Amemiya, Yuichi Suzuki, Naoto Kunii, Kensuke Kawai,
  Harushi Mori, Akira Kunimatsu, Nobuhito Saito, Shigeki Aoki, and Kuni Ohtomo.
\newblock Machine learning of dti structural brain connectomes for
  lateralization of temporal lobe epilepsy.
\newblock {\em Magnetic Resonance in Medical Sciences}, 2015.

\bibitem{bennett2019learning}
Oscar~F Bennett, Baris Kanber, Chandrashekar Hoskote, M~Jorge Cardoso,
  Sebastien Ourselin, John~S Duncan, and Gavin~P Winston.
\newblock Learning to see the invisible: A data-driven approach to finding the
  underlying patterns of abnormality in visually normal brain magnetic
  resonance images in patients with temporal lobe epilepsy.
\newblock {\em Epilepsia}, 60(12):2499--2507, 2019.

\bibitem{yang2015lateralization}
Zhengyi Yang, Jeiran Choupan, David Reutens, and Julia Hocking.
\newblock Lateralization of temporal lobe epilepsy based on resting-state
  functional magnetic resonance imaging and machine learning.
\newblock {\em Frontiers in neurology}, 6:184, 2015.

\bibitem{gazit2016probabilistic}
Tomer Gazit, Fani Andelman, Yifat Glikmann-Johnston, Tal Gonen, Aliya Solski,
  Irit Shapira-Lichter, Moran Ovadia, Svetlana Kipervasser, Miriam~Y Neufeld,
  Itzhak Fried, et~al.
\newblock Probabilistic machine learning for the evaluation of presurgical
  language dominance.
\newblock {\em Journal of neurosurgery}, 125(2):481--493, 2016.

\bibitem{Jiang2017Hemisphere}
Hui-yan JIANG, Ruo-nan LIU, Fei-fei GAO, and Yu~MIAO.
\newblock Hemisphere symmetry feature based on tensor space and recognition of
  epilepsy.
\newblock {\em Journal of Northeastern University (Natural Science)},
  38(7):923, 2017.

\bibitem{liedlgruber2019can}
Michael Liedlgruber, Kevin Butz, Yvonne H{\"o}ller, Georgi Kuchukhidze,
  Alexandra Taylor, Aljoscha Thomschevski, Ottavio Tomasi, Eugen Trinka, and
  Andreas Uhl.
\newblock Can spharm-based features from automated or manually segmented
  hippocampi distinguish between mci and tle?
\newblock In {\em Scandinavian Conference on Image Analysis}, pages 465--476.
  Springer, 2019.

\bibitem{holler2020prediction}
Yvonne H{\"o}ller, Kevin~HG Butz, Aljoscha~C Thomschewski, Elisabeth~V Schmid,
  Christoph~D Hofer, Andreas Uhl, Arne~C Bathke, Wolfgang Staffen, Raffaele
  Nardone, Fabian Schwimmbeck, et~al.
\newblock Prediction of cognitive decline in temporal lobe epilepsy and mild
  cognitive impairment by eeg, mri, and neuropsychology.
\newblock {\em Computational Intelligence and Neuroscience}, 2020, 2020.

\bibitem{Yan2018ADL}
Ming Yan, Ling Liu, Sihan Chen, and Yi~Pan.
\newblock A deep learning method for prediction of benign epilepsy with
  centrotemporal spikes.
\newblock In {\em ISBRA}, 2018.

\bibitem{zhang2021deep}
Qinming Zhang, Yi~Liao, Xiawan Wang, Teng Zhang, Jianhua Feng, Jianing Deng,
  Kexin Shi, Lin Chen, Liu Feng, Mindi Ma, et~al.
\newblock A deep learning framework for 18 f-fdg pet imaging diagnosis in
  pediatric patients with temporal lobe epilepsy.
\newblock {\em European Journal of Nuclear Medicine and Molecular Imaging},
  pages 1--10, 2021.

\bibitem{alaverdyan2018unsupervised}
Zara Alaverdyan, Jiazheng Chai, and Carole Lartizien.
\newblock Unsupervised feature learning for outlier detection with stacked
  convolutional autoencoders, siamese networks and wasserstein autoencoders:
  application to epilepsy detection.
\newblock In {\em Deep Learning in Medical Image Analysis and Multimodal
  Learning for Clinical Decision Support}, pages 210--217. Springer, 2018.

\bibitem{ghazi2016structural}
Nayereh Ghazi and Hamid Soltanian-Zadeh.
\newblock Structural connectivity of temporal lobe structures detects temporal
  lobe epilepsy.
\newblock In {\em 2016 23rd Iranian Conference on Biomedical Engineering and
  2016 1st International Iranian Conference on Biomedical Engineering (ICBME)},
  pages 30--34. IEEE, 2016.

\bibitem{udomchaiporn20163}
Akadej Udomchaiporn, Frans Coenen, Marta Garc{\'\i}a-Fi{\~n}ana, and Vanessa
  Sluming.
\newblock 3-d volume of interest based image classification.
\newblock In {\em Pacific Rim International Conference on Artificial
  Intelligence}, pages 543--555. Springer, 2016.

\bibitem{bernhardt2015magnetic}
Boris~C Bernhardt, Seok-Jun Hong, Andrea Bernasconi, and Neda Bernasconi.
\newblock Magnetic resonance imaging pattern learning in temporal lobe
  epilepsy: classification and prognostics.
\newblock {\em Annals of neurology}, 77(3):436--446, 2015.

\bibitem{samson2018deep}
Kurt Samson.
\newblock ‘deep learning’model using artificial intelligence predicts
  surgical success in intractable temporal lobe epilepsy.
\newblock {\em Neurology Today}, 18(23):50--55, 2018.

\bibitem{munsell2015evaluation}
Brent~C Munsell, Chong-Yaw Wee, Simon~S Keller, Bernd Weber, Christian Elger,
  Laura Angelica~Tomaz da~Silva, Travis Nesland, Martin Styner, Dinggang Shen,
  and Leonardo Bonilha.
\newblock Evaluation of machine learning algorithms for treatment outcome
  prediction in patients with epilepsy based on structural connectome data.
\newblock {\em Neuroimage}, 118:219--230, 2015.

\bibitem{grys2017machine}
Ben~T Grys, Dara~S Lo, Nil Sahin, Oren~Z Kraus, Quaid Morris, Charles Boone,
  and Brenda~J Andrews.
\newblock Machine learning and computer vision approaches for phenotypic
  profiling.
\newblock {\em Journal of Cell Biology}, 216(1):65--71, 2017.

\bibitem{voulodimos2018deep}
Athanasios Voulodimos, Nikolaos Doulamis, Anastasios Doulamis, and Eftychios
  Protopapadakis.
\newblock Deep learning for computer vision: A brief review.
\newblock {\em Computational intelligence and neuroscience}, 2018, 2018.

\bibitem{tellez2012review}
Jose~F T{\'e}llez-Zenteno and Lizbeth Hern{\'a}ndez-Ronquillo.
\newblock A review of the epidemiology of temporal lobe epilepsy.
\newblock {\em Epilepsy research and treatment}, 2012, 2012.

\bibitem{Goodfellow2014GenerativeAN}
Ian~J. Goodfellow, Jean Pouget-Abadie, M.~Mirza, Bing Xu, David Warde-Farley,
  Sherjil Ozair, Aaron~C. Courville, and Yoshua Bengio.
\newblock Generative adversarial nets.
\newblock In {\em NIPS}, 2014.

\bibitem{Mirza2014ConditionalGA}
M.~Mirza and Simon Osindero.
\newblock Conditional generative adversarial nets.
\newblock {\em ArXiv}, abs/1411.1784, 2014.

\bibitem{Radford2016UnsupervisedRL}
A.~Radford, Luke Metz, and Soumith Chintala.
\newblock Unsupervised representation learning with deep convolutional
  generative adversarial networks.
\newblock {\em CoRR}, abs/1511.06434, 2016.

\bibitem{Ma2020CombiningDW}
L.~Ma, R.~Shuai, Xuming Ran, Wenjia Liu, and C.~Ye.
\newblock Combining dc-gan with resnet for blood cell image classification.
\newblock {\em Medical \& Biological Engineering \& Computing}, 58:1251--1264,
  2020.

\bibitem{Oord2016ConditionalIG}
A.~Oord, Nal Kalchbrenner, Lasse Espeholt, K.~Kavukcuoglu, Oriol Vinyals, and
  A.~Graves.
\newblock Conditional image generation with pixelcnn decoders.
\newblock {\em ArXiv}, abs/1606.05328, 2016.

\bibitem{Kingma2018GlowGF}
Diederik~P. Kingma and Prafulla Dhariwal.
\newblock Glow: Generative flow with invertible 1x1 convolutions.
\newblock {\em NeurIPS}, 2018.

\bibitem{Ran2020BigeminalPV}
Xuming Ran, M.~Xu, Qi~Xu, Huihui Zhou, and Q.~Liu.
\newblock Bigeminal priors variational auto-encoder.
\newblock {\em ArXiv}, abs/2010.01819, 2020.

\bibitem{Ran2020DetectingOS}
Xuming Ran, M.~Xu, L.~Mei, Qi~Xu, and Q.~Liu.
\newblock Detecting out-of-distribution samples via variational auto-encoder
  with reliable uncertainty estimation.
\newblock {\em ArXiv}, abs/2007.08128, 2020.

\bibitem{Cohen2020ABO}
Joseph~Paul Cohen, Tianshi Cao, C.~Huang, and D.~Hui.
\newblock A benchmark of medical out of distribution detection.
\newblock {\em ArXiv}, abs/2007.04250, 2020.

\bibitem{Ren2019LikelihoodRF}
J.~Ren, Peter~J. Liu, E.~Fertig, Jasper Snoek, Ryan Poplin, Mark~A. DePristo,
  Joshua~V. Dillon, and Balaji Lakshminarayanan.
\newblock Likelihood ratios for out-of-distribution detection.
\newblock In {\em NeurIPS}, 2019.

\bibitem{zhou2020graph}
Jie Zhou, Ganqu Cui, Shengding Hu, Zhengyan Zhang, Cheng Yang, Zhiyuan Liu,
  Lifeng Wang, Changcheng Li, and Maosong Sun.
\newblock Graph neural networks: A review of methods and applications.
\newblock {\em AI Open}, 1:57--81, 2020.

\bibitem{wu2020comprehensive}
Zonghan Wu, Shirui Pan, Fengwen Chen, Guodong Long, Chengqi Zhang, and S~Yu
  Philip.
\newblock A comprehensive survey on graph neural networks.
\newblock {\em IEEE Transactions on Neural Networks and Learning Systems},
  2020.

\bibitem{kipf2016semi}
Thomas~N Kipf and Max Welling.
\newblock Semi-supervised classification with graph convolutional networks.
\newblock In {\em 5th International Conference on Learning Representations},
  2016.

\bibitem{gopinath2019graph}
Karthik Gopinath, Christian Desrosiers, and Herve Lombaert.
\newblock Graph convolutions on spectral embeddings for cortical surface
  parcellation.
\newblock {\em Medical image analysis}, 54:297--305, 2019.

\bibitem{zeng2019graph}
Runhao Zeng, Wenbing Huang, Mingkui Tan, Yu~Rong, Peilin Zhao, Junzhou Huang,
  and Chuang Gan.
\newblock Graph convolutional networks for temporal action localization.
\newblock In {\em Proceedings of the IEEE International Conference on Computer
  Vision}, pages 7094--7103, 2019.

\bibitem{song2020graph}
Xuegang Song, Feng Zhou, Alejandro~F Frangi, Jiuwen Cao, Xiaohua Xiao, Yi~Lei,
  Tianfu Wang, and Baiying Lei.
\newblock Graph convolution network with similarity awareness and adaptive
  calibration for disease-induced deterioration prediction.
\newblock {\em Medical Image Analysis}, 69:101947, 2020.

\bibitem{alaverdyan2018feature}
Zara Alaverdyan and Carole Lartizien.
\newblock Feature extraction with regularized siamese networks for outlier
  detection: application to lesion screening in medical imaging.
\newblock {\em arXiv preprint arXiv:1805.01717}, 2018.

\bibitem{zhao2017addressing}
Yijun Zhao.
\newblock {\em Addressing Bias and Subjectivity in Machine Learning}.
\newblock PhD thesis, Tufts University, 2017.

\end{thebibliography}

\newpage

\setlength{\baselineskip}{20pt}

\section*{Figure legends}

\noindent Figure 1. Epilepsy diagnosis workflow for clinicians and neuroimaging technicians. (a) Clinicians
initially treat the semeiology-diagnosed epilepsy patients with medication. If the patient is drug-resistance, the clinician will make a presurgical evaluation and multimodal images will be collected.
Then, the clinician decides the type of surgery, either radical surgery or palliative surgery, based
on whether the epilepsy is focal or general. (b) Neuroimaging technicians analyze medical data to support clinicians. They extract features from raw data via feature engineering or directly train end-to-end models for segmentation or positioning. Their ultimate tasks are computer-aided diagnosis and prognosis. \\

\noindent Figure 2.  The non-invasive multi-modal images and electrophysiology for diagnosis and prognosis of
epilepsy, including (a) T1, T2, DTI, PET images, (b) fMRI, (c) semiology video, (d) EEG. \\

\noindent Figure 3. The \re{conventional machine learning} approach. It consists of a feature engineering step to extract and select
features from single/multiple neuroimaging modalities, and a machine learning step to perform a
classification or regression task. \\

\noindent Figure 4. The \re{deep learning} approach. An example of a supervised learning model
(a) and an unsupervised learning model (b) used for neuroimaging analysis. \\

\noindent Figure 5. An example of U-NET for segmentation of brain tissues. This framework is adapted from
an open-source project \\

\newpage

\section*{Tables}

\noindent  Table 1: Representative hand-crafted features used in epilepsy


\noindent Table 2: Localization of abnormal brain regions or pathways

\noindent Table 3: \re{Tasks} of the lateralization

\noindent Table 4:  Classification of \re{epileptic} patients and \re{healthy} controls

\noindent Table 5:  \re{Prediction of prognosis in patients with epilepsy}

\end{document}